\documentclass{IEEEtran}
\usepackage{graphicx}
\usepackage{hhline}
\usepackage{amsmath,amssymb} 
\usepackage{color}
\usepackage{multirow}
\usepackage[hidelinks]{hyperref}
\usepackage{footnote}
\usepackage{algorithm}
\usepackage{algorithmic}
\usepackage{subfig} 
\usepackage{bbold}
\usepackage{accents}
\newcommand\thickbar[1]{\accentset{\rule{.4em}{.8pt}}{#1}}
\usepackage{color, colortbl}
\usepackage{xcolor}

\usepackage{hyperref}

\definecolor{Gray}{gray}{0.97}
\definecolor{LightCyan}{rgb}{0.85,1,1}

\begin{document}

\title{Hybrid Gromov-Wasserstein Embedding for Capsule Learning}


%
%
\author{Pourya~Shamsolmoali, ~\IEEEmembership{Member,~IEEE,}
              Masoumeh~Zareapoor, Swagatam Das, ~\IEEEmembership{Senior Member,~IEEE,}  Eric~Granger, ~\IEEEmembership{Member,~IEEE} and Salvador Garc\'ia             
\thanks{P.~Shamsolmoali and M.~Zareapoor are with the Dept. of Automation, Shanghai Jiao Tong University, Shanghai, China. (Emails: (pshams, mzarea)@sjtu.edu.cn).}
\thanks{S.~Das is with Electronics and Communication Sciences Unit, Indian Statistical Institute, Kolkata, India. (Email: swagatamdas19@yahoo.co.in).}
\thanks{E.~Granger is with the Dept. of Systems Engineering, \'Ecole de technologie sup\'erieure, Universit\'e du Qu\'ebec, Montreal, Canada. (Email: Eric.Granger@etsmtl.ca).}
\thanks{S.~Garc\'ia is with Dept. of Computer Science and Artificial Intelligence, University of Granada, Granada, Spain. (Email: salvagl@decsai.ugr.es).}}

\maketitle

\begin{abstract}
Capsule networks (CapsNets) aim to parse images into a hierarchy of objects, parts, and their relations using a two-step process involving part-whole transformation and hierarchical component routing. However, this hierarchical relationship modeling is computationally expensive, which has limited the wider use of CapsNet despite its potential advantages. The current state of CapsNet models primarily focuses on comparing their performance with capsule baselines, falling short of achieving the same level of proficiency as deep CNN variants in intricate tasks. To address this limitation, we present an efficient approach for learning capsules that surpasses canonical baseline models and even demonstrates superior performance compared to high-performing convolution models. Our contribution can be outlined in two aspects: firstly, we introduce a group of subcapsules onto which an input vector is projected. Subsequently, we present the Hybrid Gromov-Wasserstein framework, which initially quantifies the dissimilarity between the input and the components modeled by the subcapsules, followed by determining their alignment degree through optimal transport. This innovative mechanism capitalizes on new insights into defining alignment between the input and subcapsules, based on the similarity of their respective component distributions. This approach enhances CapsNets' capacity to learn from intricate, high-dimensional data while retaining their interpretability and hierarchical structure. Our proposed model offers two distinct advantages: (i) its lightweight nature facilitates the application of capsules to more intricate vision tasks, including object detection; (ii) it outperforms baseline approaches in these demanding tasks. Our empirical findings illustrate that Hybrid Gromov-Wasserstein Capsules (HGWCapsules) exhibit enhanced robustness against affine transformations, scale effectively to larger datasets, and surpass CNN and CapsNet models across various vision tasks.
\end{abstract}


\section{Introduction}
Convolutional Neural Networks (CNNs) were devised to extract significant features and attributes from images, mirroring the principles of the neuronal receptive fields in the human visual cortex. This is achieved through a collection of small kernels that identify distinct patterns \cite{li2021survey}. CNNs have achieved magnanimous success in many computer vision-related applications as they can learn semantic feature representations from input data. However, capturing the hierarchy of spatial relationships among different parts of an object is challenging for CNNs because of their poor scaling properties concerning large receptive fields \cite{hinton2021represent}. Indeed, CNNs have a fixed-size receptive field, which limits their ability to capture spatial relationships across multiple scales. To illustrate, let's consider an object such as a face. While a CNN can identify its distinct features like the mouth and nose, it fails to detect the precise locations of these features. The network only recognizes their presence without taking into account their respective positions. However, this parsing mechanism with hierarchical relationship modeling is in line with capsule networks (CapsNets) \cite{hinton2018matrix}\cite{kosiorek2019stacked}\cite{sabour2017dynamic}. In other words, CapsNets are designed to present objects as a set of parts and their relationships, while such a complex structure is hard to impose in a standard CNN \cite{hinton2021represent}. 
CapsNets create this hierarchical representation through a two-phase process. The first phase is a part-whole transformation, aiming to capture different image features and parts. The second phase, called hierarchical component routing, learns the relationships between these features and parts to create a hierarchical representation of the input. The dynamic routing mechanism in CapsNets plays a crucial role in accurately sorting out the hierarchy of the parts and determining which part belongs to which object \cite{hinton2021represent}\cite{sabour2017dynamic}. CapsNet has unique advantages over CNN-based models due to its ability to model the spatial relationships between objects and parts, which can be particularly useful in tasks such as image classification \cite{hinton2018matrix}\cite{sabour2017dynamic}, object detection and segmentation \cite{lalonde2021deformable}\cite{lalonde2018capsules}, video analysis \cite{duarte2022routing}, image generation \cite{zhao20223dpointcaps++}, graph learning \cite{yang2022ncgnn}, and also performed well in face-part detection \cite{yu2022hp}\cite{yu2023graphics}. 

Due to several limitations, the current CapsNets are less widely applicable than CNNs. First, the pairwise relationships between capsules in consecutive layers often rely on global information and may not effectively capture local interactions \cite{garau2022interpretable}. In other words, while capsules are effective in routing information from different locations in an image, they cannot capture effective low-level parts in the image, which may limit their effectiveness on complex tasks \cite{sabour2021unsupervised}. Second, the pairwise modeling between capsules of two layers increases the number of parameters, and this design decreases the generalizability, leading to overfitting during training \cite{mitterreiter2023capsule}\cite{chen2022tabcaps}. Third, routing-by-agreement involves iterative operations between capsules, which can be computationally expensive, limiting its use to only small-scale datasets \cite{hinton2021represent}\cite{mitterreiter2023capsule}\cite{paik2019capsule}. Indeed, several works have empirically reported scaling issues in CaspNets, highlighting their limitations when applied to larger datasets or more complex tasks. Specifically, the studies by \cite{mitterreiter2023capsule}\cite{paik2019capsule} have provided valuable insights into these scaling challenges.
Choi et al. \cite{choi2019attention}  introduced the attention concept into capsule routing, employing a feed-forward operation instead of iterative updates. However, this design of routing, while computationally more efficient, achieved only 87.54\% accuracy on CIFAR-10. Other works \cite{pucci2021self}\cite{hahn2019self}, have explored incorporating self-attention mechanisms into CapsNets. These methods aim to reduce the computational complexity of the routing mechanism by calculating attention weights for each capsule based on its interactions with other capsules in the network. The attention weights are then used to adjust each capsule's contribution to the final capsule output. While they show promise on small-scale tasks, their performance on more complex tasks still needs to improve.

\begin{figure}[t]
\centering
\includegraphics[height=4cm]{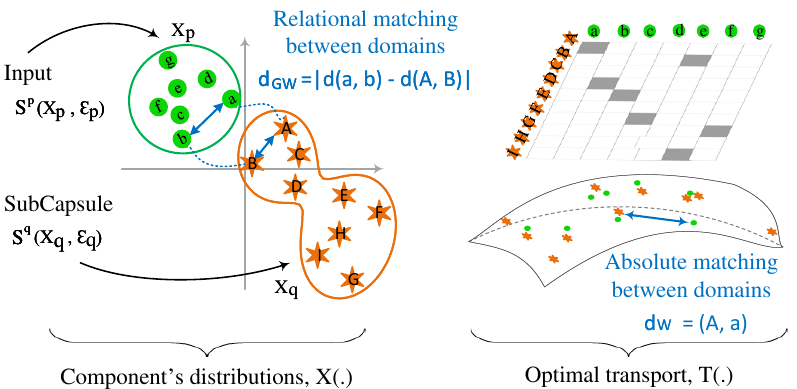} 
\caption{The learning strategy of the Hybrid Gromov-Wasserstein Capsules is based on the relevance of input to a group of subcapsules. We interpret input and subcapsule as two domains that define data distributions. The proposed mechanism uses the Gromov-Wasserstein distance to measure the dissimilarity between the input and each subcapsule (left) and find their level of alignment according to the optimal transport (right). }
\label{fig:1}
\end{figure}

Differing from the existing CapsNets, we propose a new framework for capsules that creates capsules by learning a group of subcapsules for different classes. In particular, we propose a Hybrid Gromov-Wasserstein mechanism that measures the discrepancy between the input and subcapsules and finds their alignment score based on the optimal transport (see Fig.~\ref{fig:1}). This relational regularizer allows CapsNets to learn from noisy, high-dimensional input effectively and serves as an efficient alternative to existing routing strategies.
Our capsule is a lightweight model that can be successfully incorporated into existing deep-CNN approaches with a significant improvement in the error rate. To the best of our knowledge, this is the only framework for CapsNets that allows adapting the capsule distributions to another capsule and achieves comparable performance as the deep CNN approaches (e.g., ResNet-18), with fewer parameters; e.g., our model achieves 80.12\% test accuracy on CIFAR-100, and 96.78\% on CIFAR-10 with 30\%$\sim$40\% few parameters. \\

\begin{figure*}[t]
\centering
\includegraphics[height=4.6cm]{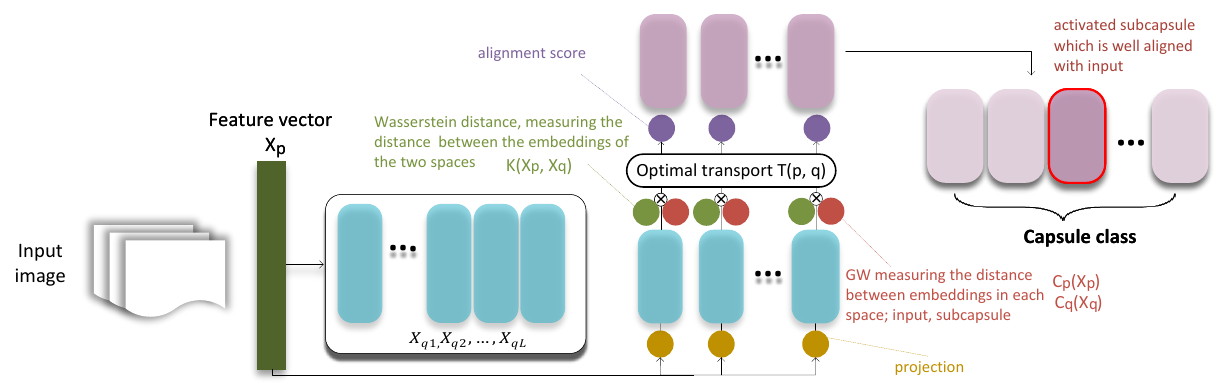} 
\caption{Illustration of our model. The feature vector $X_p$ is generated by a backbone network from an input image. Given its ground truth label $y\in\{1, ..., L\}$, we would create $L$ subcapsules, each corresponding to one of $L$ classes. Then the input is transported onto each of subcapsule $X_q$, yielding the optimal transport plan $T(X_p, X_q)$ to find the mapping between them (the green and red circle indicate the GW and Wasserstein distance between an input and each subcapsule). During training, our model learns to find the alignment between the input and the corresponding subcapsule for the correct class.}
\label{fig2}
\end{figure*}

{\it Contributions} This study introduces an innovative capsule derived from the Hybrid Gromov-Wasserstein framework. Agreement-based models such as dynamic routing require iterative computations among capsules for consensus, which can be computationally expensive  \cite{mitterreiter2023capsule}. 
However, we adopt a different learning approach. We calculate a distance measure between the input and each subcapsule, designating the subcapsule with the minimal distance as the higher-level capsule (deeper capsule). Each subcapsule is generated for every input feature vector class, capturing potential variations in object components for each specific class. This routing based on distance metric obviates the necessity for iterative coefficient updates, thereby enhancing training efficiency and scalability. Our approach combines Gromov-Wasserstein (GW) and Wasserstein metrics to achieve this. GW is a mathematical framework that measures relational dissimilarity between two metric spaces, even if they have different underlying geometric structures \cite{peyre2016gromov}. For example, it can be used to compare images with different resolutions or orientations. On the other hand, Wasserstein distance computes the minimum cost of transporting distributions from one space to another. The key advantage of our proposed framework is its ability to efficiently compute the alignment score between an input and a subcapsule using Sinkhorn's algorithm \cite{cuturi2013sinkhorn}. Fig.~\ref{fig2} illustrates our proposed model. Our contributions throughout the paper can be summarized in the following way.
\begin{itemize}
\item We introduce a new capsule learning method, where the capsules are created according to the relevance of an input feature vector to a group of subcapsules. The HGW framework is proposed for this purpose.
\item Routing-based distance does not require iterative computations. They directly evaluate the alignment between subcapsules and the input, leading to reduced computational overhead. Subcapsules strongly related to the input receive the shortest distance, while those with little or no connection are removed. 
\item The performance of the HGWCapsule is assessed across a range of demanding datasets, including COCO, ImageNet, CIFAR-100, SmallNORB, SVHN, Pascal VOC, and ISIC-2018. Notably, our proposed model surpasses the CapsNet and CNN baselines while utilizing fewer parameters. Furthermore, the model exhibits remarkable scalability, demonstrating effectiveness even on larger datasets within various vision tasks.
\end{itemize}
 \color{black}

 {\it Organization}. We discuss the related works in Sec.~\ref{related}. Sec.\ref{sec3a} introduces the concept of subcapsules, \ref{sec3c} defines the notations used in this paper, and \ref{sec3b} presents the core framework of our model. Experimental results and analysis are provided in Sec.~\ref{expriment}. Finally, in Sec.~\ref{conc} we end with the conclusion and ideas for future work.
\section{Related Work}
\label{related}
Given the vast literature on capsule models and Wasserstein distances, we only discuss the most recent related works here. \\
\vspace{-4pt}

\subsubsection{Capsule Networks} 
The notion of capsules was initially introduced by Hinton et al. \cite{hinton2011transforming}. They highlighted the limitations of traditional CNNs in achieving viewpoint invariance and proposed capsules as a more intricate structure to overcome these constraints. Subsequently, Sabour et al. \cite{sabour2017dynamic} delved into the concept of capsules and introduced CapsNet by incorporating the dynamic routing (DR) algorithm. This algorithm serves as a core element of CapsNets, facilitating the construction of capsule parse trees from images. CapsNet demonstrated superior performance compared to CNN baselines on MNIST, exhibiting fewer parameters and heightened resistance to affine transformations. This success ignited substantial interest, leading to subsequent research endeavors to enhance and expand the CapsNet architecture.
In the DR process, all information on lower-layer capsules is used to build the higher-layer capsules. Thus, many unimportant capsules will be created, leading to an expensive design. Several subsequent works have attempted to produce more efficient and scalable routing algorithms, including Expectation-maximization (EM) routing \cite{hinton2018matrix}, routing-based attention \cite{hahn2019self}\cite{mazzia2021efficient}\cite{tsai2020capsules}, and entropy routing \cite{wang2018optimization}\cite{renzulli2022rem}. Considering that rich information can be stored in capsules, they have been extensively used in various domains and problems, including graph networks \cite{fu2022tempcaps}, medical \cite{lalonde2018capsules}, 3D objects from point clouds \cite{zhao20223dpointcaps++}, Cyber attack \cite{mahdavifar2023explainable}, and generative models \cite{edraki2020subspace}. Kosiorek et al. \cite{kosiorek2019stacked} proposed an unsupervised capsule autoencoder (SCAE) for object representation using visualizable templates, showing promising results on simple image classification datasets. Further, Yu et al. \cite{yu2022hp} extended SCAE to discover hierarchical face parts and their relationships from unlabeled input images, but it has limitations with faces having small poses.

Subspace capsule networks (SCN) \cite{edraki2020subspace} create scalable capsules through grouping subspace capsules rather than merely grouping neurons. Unlike CapsNet \cite{sabour2017dynamic}, where all low-level capsules contributed to constructing the higher-level capsules, in SCN, the low-level capsules will look for an agreement among themselves; if there is a strong agreement, these capsules can be activated as the high-level capsules. The model's performance was evaluated on the CIFAR-10, achieving an accuracy of 89.64\%.
Jia et al. \cite{jia2020capsnet} proposed a diverse and enhanced capsule network (DE-CapsNet) based on the residual blocks and position-wise dot product operation to improve the capsule's performance on complex datasets. DE-CapsNet achieved 94.25\% and 91.98\% accuracy on the Fashion-MNIST and CIFAR-10 datasets, respectively.
CVAE-Capsule \cite{guo2021conditional} is proposed for open set recognition, and the model achieved 95.6\% and 83.5\% test accuracy on SVHN and CIFAR-10, respectively. 
To extract and encode salient features from complex data, Dense capsule networks (DenseCaps) \cite{sun2022novel} were proposed. DenseCaps achieved an accuracy of 99.57\%, 95.78\%, and 89.31\% on MNIST, SVHN, and CIFAR-10 datasets, respectively. Garau et al. \cite{garau2022interpretable} proposed a novel capsule framework that represents interpretable part-whole relationships in data, and they evaluated its performance on simple image classification tasks. 

Lalonde et al. \cite{lalonde2018capsules} proposed a deconvolution capsule network for medical image segmentation, surpassing UNet \cite{ronneberger2015u} by 0.03\% in accuracy. DeformCapsule \cite{lalonde2021deformable} adapted the Squeeze-and-Excitation network \cite{hu2018squeeze} and introduced a new routing mechanism to maximize capsule agreement without iteration.
IDPACapsule \cite{tsai2020capsules} is based on sequential iterative routing. This model with a ResNet backbone achieved an accuracy of 95.14\% and 78.02\% on CIFAR-10 and CIFAR-100, respectively (compared to 95.11\% and 77.53\% of the original backbone). 
While these methods have found successful applications to many problems, training capsules can be challenging and require many hyperparameters. Most CapsNets perform well on simple datasets like CIFAR-10, but their performance degrades significantly on more complex tasks as the complexity of the problem increases. \\

\vspace{-2pt}


\subsubsection{The Gromov-Wasserstein distance and its application} 
Wasserstein distance, also known as Optimal Transport (OT) is a powerful approach for comparing two distributions that both rely on the same space \cite{villani2008optimal}. It quantifies the distance between distributions by computing the minimum cost of transforming one distribution into the other. Unlike other metrics such as Euclidean or cosine distance, Wasserstein distance can capture subtle differences between distributions  \cite{peyre2017computational}. It is a stable metric, meaning small changes in the input data result in small changes in the distance. This is important in vision tasks where there can be variations in the data due to lighting conditions, camera angles, or other factors \cite{fuchs2020wasserstein}. 
However, when dealing with distributions in different metric spaces, the Wasserstein distance may not be adequate and a more complex framework, such as the Gromov-Wasserstein (GW) \cite{memoli2014gromov} model, is needed. The GW distance compares two samples located in different metric spaces by considering their geometric structures and relationships. 
Due to its ability to use the geometric characteristic of data, GW models have been used in many applications, e.g., natural language processing \cite{alvarez2018gromov}, computer graphics \cite{peyre2016gromov}\cite{xu2019gromov}\cite{xu2022representing}, and computer vision \cite{nakagawa2022gromov}. 
In \cite{bunne2019learning}, a generative method is proposed to learn a pair of distributions lying in spaces
of different dimensions, which uses the GW metric as the objective function. In \cite{pmlr-v119-xu20e} and \cite{titouan2020co}, the authors adapted the GW distance to build VAE for graph classification and for domain adaptation over different spaces, respectively. 
To enhance the efficiency of the GW, Peyr\'e et al. \cite{peyre2016gromov} and Le et al. \cite{le2022entropic} introduced regularization based on the entropy of the transportation plan, which can help to avoid overfitting and improve the generalization ability of the model.   

Another line of research \cite{vincent2021semi} relaxes the GW discrepancy for quantifying the similarities between two or more graphs. This framework combines the GW model with the Sinkhorn algorithm, making it especially effective for large and complex networks. 
In image processing, Liu et al. \cite{liu2022sparsity} introduced the sparse optimal transport model to efficiently assign many pixels in an image to a smaller set of image features. The method uses a sparsity constraint to reduce the number of variables that need to be optimized, leading to a more efficient and compact representation of the problem with reduced computational cost. 
Kolouri et al. \cite{kolouri2020wasserstein} introduced Wasserstein embedding for graph learning, representing graphs as probability distributions over sub-graphs and using OT to compare the distributions. The method performed well on graph classification and molecular property-prediction benchmarks. However, it needs more consideration of node features, making it insufficient for problems where node features are crucial. A similar work by Mialon et al. \cite{mialon2020trainable} is proposed, where the Wasserstein embedding is used as a pooling operator for learning from sets of features. Their model outperformed standard attention mechanisms in terms of accuracy and speed. However, the Wasserstein distance only considers the distance between the distributions and not their spatial relationship \cite{NEURIPS2021_4990974d}. 
Unlike Wasserstein or GW distances that focus respectively on the distribution of features or their spatial relationships, we wish to offer a model that jointly exploits both information and is consequently called the Hybrid Gromov-Wasserstein framework. By leveraging the Wasserstein distance for comparing low-level features (e.g., color and texture) and the GW distance for comparing higher-level features (e.g., shape and structure), our model effectively compares complex distributions and finds their best correspondence. This hybrid approach allows capsules to be employed in more complicated tasks while providing better or equivalent performance to CNN-based models.

\section{Hybrid Gromov-Wasserstein for Capsules}
\label{sec3a}
Let $x\in R^d$ be a feature vector generated from an input image using a backbone network to represent the input entity. With its ground truth label $y\in\{1,..., L\}$, we aim to learn a collection of subcapsules $\{X_{q_1},..., X_{q_L}\}$, each of which is associated with one of the classes $L$; i.e., each subcapsule represents the potential variations of object parts for a specific class. Then, our model learns to find the alignment score between the input and the corresponding subcapsule for the correct class. Existing capsule models \cite{duarte2022routing}\cite{edraki2020subspace} apply many transformation matrices to each of the lower-level capsules to generate a set of predictions. These predictions are then fed into the routing-by-agreement process to construct deeper capsules. Consequently, they must train several parameters to learn the pairwise relationships between capsules in the subsequent layers. In this paper, we instead consider that the subcapsules should be activated if they align well with the input. Alignment scores will be obtained by computing the proposed HGW between the input and the subcapsule sets. Once the correct subcapsule has been identified, it is used to construct the higher-level capsules. Therefore, this output consists of the pose vector of the object, which encodes the spatial information about the object, and a probability vector that represents the object belonging to each of the L classes. In this design, the number of alignments obtained for a given layer depends on the number of subcapsules in that layer. 

\subsection{ Definition and Properties} 
\label{sec3c}
We use the following notations throughout the paper. We treat the input and subcapsule as two individual domains that are denoted by $X_p$ and $X_q$ with dimensional space $V_p$, and $V_q$, respectively (where $V_p\geq V_q$). Each domain is comprised of a set of embeddings and a set of possible interactions,  $\mathcal{E}_s=\{(u_i, u_j, w_{ij})~\vert~(u_i, u_j) \in V_\theta \}$ for ($\theta$ = $p$ or $q$), and $w_{ij}$ counts the possible interactions. Given an input feature vector and subcapsules, we consider finding their alignment score by i) computing the similar embeddings of the two domains (e.g., $X_p=[x_i^p]\in R^{d\times V_p}$ and $X_q=[x_i^q]\in R^{d\times V_q}$); (ii) finding the degree of alignment (correspondence) between the input and each subcapsule. These objectives are unified using the proposed Hybrid Gromov-Wasserstein model, which performs significantly better for complex datasets in practice.  
\vspace{-5pt}

\subsection{The Gromov-Wasserstein between spaces}\label{sec4}
The GW \cite{memoli2014gromov} metric measures the discrepancy between probability distributions supported on different spaces. It involves distance matrices $C_p=[c^p_{i j}]\in R^{V_p\times V_q}$ and $C_q=[c^q_{i' j'}]\in R^{V_q \times V_q}$, representing the distances within the input domain and the subcapsule domain, respectively. These distance matrices are constructed based on the interaction sets $\mathcal{E}$, and their elements $c_{ij}$ is a function of $w_{ij}$. Thus, the GW metric between these two domains is given by
\begin{equation}
\begin{array}{lr}
\underset{T\in \Gamma(\mu_p, \mu_q)}{\min}\sum_{i, j, i', j'} T_{i i'} T_{j j'}~L(c^p_{i j}, c^q_{i' j'}), \\
\hfil \hfil =\underset{T\in \Gamma(\mu_p, \mu_q)}{\min} \langle L(C_p, C_q, T), T \rangle. 
\end{array}
\label{eq:8}
\end{equation} 
where $\Gamma$ is the transportation polytope that can be interpreted as the set of probabilistic distributions between $\mu_p$ and $\mu_q$;  $\Gamma(\mu_p, \mu_q):=\{T\in R^{V_p \times V_q};~T \mathbb{1}_{ V_q}=\mu_p~\text{and}~T^T \mathbb{1}_{ V_p}=\mu_q\}$. Loss function $L(\cdot, \cdot)$ is computed using KL-divergence $L(a, b)=a~\text{log}\frac{a}{b}-a+b$. Accordingly, $\langle \cdot, \cdot\rangle$ is the Frobenius dot product of two matrices. Moreover, $L(C_p, C_q, T)=[L_{j j'}]\in R^{ V_p\times  V_q}$, where its element $L_{j j'}=\sum_{i,i'} L(c^p_{i j},~c^q_{i' j'}) T_{i' i}$. $T$ is the optimal transport between the components and its elements $T_{i j}$ indicates the probability that $x_i\in V_p$ is relevant to $x_j\in V_q$. 
This formulation of GW for learning capsules poses the following key problems:
\begin{itemize}
\item The feature representations in the input and subcapsule domains are different, requiring a method to establish correspondence and reduce domain discrepancy.
\item The underlying space of subcapsule is a manifold, while the input is in a vector space, the obtained interactions between these two can be noisy, leading to an unreliable distance matrix.
\item The GW model only captures the structural information in the transportation problem. At the same time, we are interested in jointly exploiting features and structural information in a unified framework to achieve the correspondence between the input and subcapsules.
\end{itemize}
Therefore, we impose an optimal transport between the embeddings of different domains to perform their correspondence/alignment efficiently. We will formally define our model in the following section.

\subsection{Proposed model} 
\label{sec3b}
We propose the Hybrid Gromov-Wasserstein (HGW) framework to obtain meaningful mappings and correspondence between the input and subcapsules. 
To achieve this, we define the capsules of a layer $l$ as an output of our alignment scheme produced by the preceding layer $(l-1)$. We present our new formulation and its properties as 
\begin{equation}
\begin{split}
\min_{X_p,  X_q} \min_{T\in \Gamma(\mu_p,\mu_q)}\underbrace{\langle L(C_p(X_p), C_q(X_q), T), T}_{Gromov-Wasserstein~distance}\rangle + \\ \underbrace{\beta\langle K(X_p, X_q), T\rangle}_{Wasserstein~distance} +  \underbrace{E(X_p, X_q)}_{Regularizer}.
 \label{eq:21}
\end{split}
\end{equation}

The first term represents the GW metric, which measures the relational dissimilarity between the input and subcapsule spaces by comparing the difference between the pairs of samples from the two distributions. More details are discussed in Eq.~(\ref{eq:8}). The distance matrix is computed by $C_\theta (X_\theta)=(1-\beta)C_\theta + \beta  K (X_\theta, X_\theta)$, for $\theta=p$ or $q$ that combines both the information from the observed ($C_\theta$), and embeddings ($K(X_\theta, X_\theta)$) that are defined as follows:
\begin{equation}
K (X_\theta, X_\theta)=[k(x_i^\theta, x_j^\theta)] ~~~~~ \text{for} ~~ \theta = p ~\text{or} ~ q,
\end{equation}
where its elements $[k(x_i^\theta, x_j^\theta)]\in R^{V_\theta \times V_\theta}$ is a function computing the distance between the embeddings in each domain or space, the intuition behind this is that the observed data may not fully capture the underlying features of the data, especially if the data is high-dimensional or nonlinear. In contrast, embeddings are low-dimensional representations of the data that capture its underlying information in a more meaningful way. The parameter $\beta\in[0, 1]$ controls the relative weight of the two distance metrics in the overall distance. When $\beta=0$, only the predefined distance metric $C_\theta$ is used, and when $\beta=1$, only the embedding $K(X_\theta,X_\theta)$ is used. When 0 $<\beta<$  1, the distance matrix is a weighted combination of the two distance metrics.

The second term, measures the absolute dissimilarity between the input embeddings and subcapsules. In particular, the Wasserstein distance measures the amount of "work" required to transform one distribution into another, where the "work" is defined as the distance between the elements in the distribution \cite{peyre2017computational}. Using this term, we can efficiently find an optimal mapping that connects the embeddings of the two spaces, ensuring an accurate correspondence between the samples in one space and their counterparts in the other space \cite{kolouri2020wasserstein}. The distance matrix is obtained based on the embeddings $K(X_p, X_q)=[k(x_i^p, x_j^q)]\in R^{V_p \times V_q}$, and its contribution is controlled by the the same hyperparameter $\beta\in [0, 1]$.  However, one advantage of the Wasserstein distance is that it is more robust to noise and outliers than the GW distance,  because it is based on comparing probability distributions directly, rather than comparing their embeddings in metric spaces \cite{peyre2017computational}. Another advantage is that it is computationally more efficient and easier to implement \cite{memoli2011gromov}. 
\begin{algorithm}[t]
	\caption{Training of the proposed capsule. Our algorithm returns the activation and poses of the capsules in layer $l+1$ based on the alignment score in layer $l$; the capsules are activated for the upper layers by finding the alignment between the input and subcapsule sets.}
	\label{algo:2}
     {\bf Input:} $\{C_p, \mu_p\}, \{C_q, \mu_q\}$, $\lambda$, $\epsilon$, number of outer-inner iterations \{$\mathcal{U}, \mathcal{V}$\}.\\
 {\bf Output:} $X_p$, $X_q$ and $\thickbar{T}$.\\
Initialize $X_p^{(0)}$, $X_q^{(0)}$, $\thickbar{T}^{(0)}=\mu_p \mu_q^T$. \\
 {\bf For} $\mathcal{U}=0$ to U-1 \par  
\hspace{5mm} Set $\thickbar{\beta}_\mathcal{U} = \frac{\mathcal{U}}{U}$.\par
\hspace{5mm} {\bf For} $\mathcal{V}=0$ to V-1 \par
\hspace{8mm} Update optimal transport $\thickbar{T}^{(\mathcal{U}+1)}$ \hfill  via (Eq. \ref{eq:25}). \par
\hspace{5mm} Obtain $X_p^{(\mathcal{U}+1)}, X_q^{(\mathcal{U}+1)}$ \hfill via (Eq. \ref{eq:27}). \\
        $X_p= X_p^{(\mathcal{U})}$, $X_q = X_q^{(\mathcal{U})}$, $\thickbar{T}=\thickbar{T}^{(\mathcal{U})}$. \\
    $\setminus\setminus$  we set the correspondence/alignment $\mathcal{A} = \emptyset$ \\
       {\bf For} $v_i\in V_p$ \par
\hspace{5mm} $j=argmax_j$ $\thickbar{T}_{ij}$. $\mathcal{A}= \mathcal{A}\cup\{(v_i\in V_p, v_j\in V_q)\}$. 
\end{algorithm}
The combination of these metrics has two perspectives. First, the optimal transport indicates the alignment between two spaces \cite{memoli2011gromov}\cite{xu2019scalable}. Furthermore, both metrics share a standard mathematical structure to compare the distributions computationally efficiently, making it possible to use the method for large datasets. The sparse nature of the resulting optimal transport can lead to overfitting or incorrect matching of samples \cite{gu2022keypoint}. By regularizing the embeddings, we can improve the quality and consistency of the distance metric by reducing the impact of noise or distortions in the embedding process. We further use the regularization $E$ based on $C_p$ and $C_q$, then obtain
\begin{equation}
E(X_p, X_q)=\lambda \sum_{\theta = p,q} L(K(X_\theta, X_\theta), C_\theta).
\label{reg}
\end{equation}
where loss function $L(\cdot,\cdot)$ is defined in section (\ref{sec4}). The weighting factor or regularization parameter $\lambda$  controls the strength of the penalty and helps to balance the trade-off between fitting the training data, set to 10 in our implementation. The above processes are unified in the same framework for capsule learning. Using the GW discrepancy term, we obtain the embedding-based distance matrices that can reduce the noise and irrelevant features in the data-driven distance matrices, thereby improving robustness. Moreover, we compute the Wasserstein distance between the embeddings of input and subcapsules to find the best mapping (correspondence) between them through the optimal transport, in which the optimal transport values determine which input and subcapsule pairs should be close to each other.

\section{Optimization and algorithmic solution}
\label{sec5}
As can be seen, (\ref{eq:21}) is a quadratic problem; thus directly learning the optimal transport is difficult. We address this problem by alternatively learning the optimal transport and the embeddings through a Proximal Gradient (PG) algorithm \cite{xu2022representing}\cite{xu2019scalable}. Based on the inner-outer iteration loops, in the $\mathcal{U}^{th}$-outer iteration for the given input and subcapsule embeddings $X_p^{(\mathcal{U})}$ and $X_q^{(\mathcal{U})}$, the corresponding update of Eq.(\ref{eq:21}) can be expressed as   
\begin{equation}
\begin{array}{lr}
& \underset{T\in \Gamma(\mu_p,\mu_q)}{\min}\langle L(C_p (X_p^{(\mathcal{U})}), C_q (X_q^{(\mathcal{U})}, T), T\rangle + \\ & \beta \langle K (X_p^{(\mathcal{U})}, X_q^{(\mathcal{U})}), T\rangle.
\end{array}
\label{eq:24}
\end{equation}

We remark that this problem still holds nonconvexity due to the quadratic term $\langle L(C_p (X_p^{(\mathcal{U})}), C_q (X_q^{(\mathcal{U})}, T), T\rangle$. Consequently, we propose to solve (\ref{eq:24}) using a PG algorithm that can benefit from alternatively learning the optimal transport and the embedding. Thus, in the $\mathcal{V}^{th}$-inner iteration, the optimal transport will be updated via: 
\begin{equation}
\begin{array}{lr}
&\underset{T\in \Gamma(\mu_p,\mu_q)}{\min}\langle L(C_p (X_p^{(\mathcal{U})}), C_q (X_q^{(\mathcal{U})}), T), T\rangle + \\
&\beta \langle K (X_p^{(\mathcal{U})}, X_q^{(\mathcal{U})}), T\rangle + \epsilon \underbrace{KL(T\Vert T^{(\mathcal{V})})}_{proximal~term},
\end{array}
\label{eq:25}
\end{equation}
where $\epsilon$ is the regularization parameter and $KL(T\Vert T^{(\mathcal{V})})$, is uses as a regularizer. The purpose of this regularization term is to impose constraints on the transportation plan during the optimization process, which is computed by $\sum_{ij} T_{ij}\log \frac{T_{ij}}{T_{ij}^{(\mathcal{V})}}-T_{ij} + T_{ij}^{(\mathcal{V})}$ \cite{shamsolmoali2023vtae}. Following \cite{peyre2016gromov}, this problem can be solved via a projected gradient descent method, where the projection relies on the KL metric divergence metric. In our implementation, we set $\epsilon=0.001$. \\

\noindent {\bf Proposition 1.} {\it In the special case when the learning rate is $1/\epsilon$, the Eq. \ref{eq:25} uses projected gradient descent to solve the optimal transport problem via an entropy regularizer.} \\

\noindent As shown in \cite{peyre2016gromov} and \cite{benamou2015iterative}, the projection is the solution to the regularized transport problem. It involves finding the transportation plan that minimizes the KL-divergence between $X_p$ and $X_q$ subject to a regularization term (the proof is given in the \hyperref[proof]{Appendix}). This regularization can be added to the transport problem in the following way to ensure that the solution is stable. 

\begin{equation}
\begin{aligned}
\underset{T\in \Gamma(\mu_p, \mu_q)}{\min}\langle C^{(\mathcal{U},\mathcal{V})}, T\rangle-\epsilon H(T), \\
C^{(\mathcal{U},\mathcal{V})}=L(C_p, C_q, T^{(\mathcal{V})})+\beta K(X_p^{\mathcal{U}}, X_q^{\mathcal{U}})+\epsilon.
\label{eq:25a}
\end{aligned}
\end{equation}

The Sinkhorn's matrix scaling algorithm can solve this problem \cite{cuturi2013sinkhorn} with linear convergence, and $H(T) =-\sum_{ij} T_{ij} {\text ln} T_{ij}$.
This regularization yields favorable computational properties since it defines a convex algorithm that involves a Sinkhorn projection  \cite{cuturi2013sinkhorn} by only performing matrix-vector multiplications. The problem (\ref{eq:24}) is broken down into several steps. Each step (\ref{eq:25}) can be solved through the projected gradient descent, which provides a solution for a regularized optimal transport problem (\ref{eq:25a}). 
Furthermore, the effect of embeddings $X_p$ and $X_q$ on the GW and Wasserstein discrepancies is controlled by parameter $\beta$. Since these embeddings are randomly initialized. As a result, they are unreliable at the beginning of the training process. We address this numerical instability by drastically growing $\beta$ w.r.t. the number of outer iterations. This assumption can be approximated by $\thickbar{\beta}_\mathcal{U}=\frac{\mathcal{U}}{U}$, where $U$ is the expected (maximum) number of outer iterations, in the $\mathcal{U}$-th iteration.  Thus, given the optimal transport $\thickbar{T}^{(\mathcal{U})}$, the embeddings will be updated through the following optimization problem 
\begin{equation}
\min_{X_p, X_q} \thickbar{\beta}_\mathcal{U}\langle K(X_p, X_q), \thickbar{T}^{(\mathcal{U})}\rangle + \lambda E(X_p, X_q).
\label{eq:27}
\end{equation}
Stochastic gradient descent is used to solve this problem.

\subsection{Distance matrix}
\label{sec:act}
The distance matrix is an essential component of our proposed framework. A short distance for a deeper capsule indicates that the input feature vector is well aligned with the subcapsule, i.e., the input contains the entities/features represented by the subcapsules. We count the interactions between input and subcapsules based on Zipf's law \cite{powers1998applications}, which treats the counts as the weights of interactions. Following this perspective, we define the element of the distance matrix as
\begin{equation}
c_{ij}= 
\begin{cases}
\frac{1}{w_{ij}+1} ~~~~ (u_i, u_j)\in \mathcal{E}_\theta; ~ \forall \theta=p,q \\
1 \; \;  ~~~~~~~~~ (u_i, u_j)\not\in \mathcal{E}_\theta; ~ \forall \theta=p, q. \\
\end{cases}
\label{eq:18}
\end{equation}

where $w_{ij}$ is the count of interactions between the input and each subcapsule. Indeed, the magnitude of $c_{ij}$ indicates the degree of alignment of input with the subcapsule. Thus, the subcapsules will be activated only when their embeddings are in the right relationship with the input (i.e., the input is a part of the entity modeled by a subcapsule), or remove the subcapsule if its alignment score is close to one. This property makes our capsule easily scalable to large network architectures and large datasets. Algorithm \ref{algo:2} presents the proposed learning strategy for capsule networks.

\subsection{Computation complexity}
Beyond the significant performance improvement, one of the advantages of the proposed capsule model is its computational efficiency. Learning optimal transport is computationally hard in the general case due to the tensor-matrix multiplication $L(C_p, C_q, T)$. Peyr\'e et al. \cite{peyre2016gromov} suggested an efficient way to compute the loss to provide a successful learning process. For functions $(f_1, f_2, h_1, h_2)$ the loss $L(a, b)$ can be defined as $L(a, b)=f_1(a)+f_2(b)-h_1(a)h_2(a)$. Therefore for any $T\in \Gamma(\mu_p, \mu_q)$, the computation problem can be rewritten as
\begin{equation}
\begin{aligned}
 L(C_p, C_q, T)= \\
f_1(C_p)\mu_p 1^T_{V_q} + 1_{V_p}\mu_q^T f_2(C_q)^T - h_1(C_p) T h_2(C_q)^T.
\label{eq:10}
\end{aligned}
\end{equation}

The sparsity of the optimal transport causes the complexity of L to be $\mathcal{O}(M^3)$, where $M=\{V_p, V_q\}$. The complexity of the distance matrix $K(X_p, X_q)$ is $\mathcal{O}(M^2 D)$, for D-dimensional embedding. Additionally, to optimize the transport plan, we used the proximal point method \cite{xu2018distilled}, which involves running a Sinkhorn-Knopp projection in each inner iteration. Thus, the overall complexity of the learning model will be $\mathcal{O}(U(M^2D + V M^3))$, where $U$ is the number of outer iterations and $V$ is the number of inner iterations. This process only requires matrix multiplication and is easy to implement.


\setlength{\tabcolsep}{2.8pt}
\begin{table*}[t]
\centering
\caption{Affine transformation experiments on MNIST (values are in \%)}
\label{tab:1}
\begin{tabular}{c|ccccclc|ccccc} \cline{1-6}\cline{8-13} 
\multirow{2}{*}{Method}& \multicolumn{5}{c}{\textbf{Training on Untransformed Data }} & &\multirow{2}{*}{Method}& \multicolumn{5}{c}{\textbf{Training on Transformed Data} (2, $30^\circ$)} \\ \cline{2-6}\cline{9-13}
 & (0, $0^\circ$) & \hfil (2, $30^\circ$) & \hfil (2, $60^\circ$) & \hfil (2, $90^\circ$) &  (2, $180^\circ$) & &   & (0, $0^\circ$) & \hfil (2, $30^\circ$) & \hfil (2, $60^\circ$) & \hfil  (2, $90^\circ$) &   (2, $180^\circ$) \\ \cline{1-6}\cline{8-13}
SCN & 99.38 & 92.76 & 72.55 & 59.31 & {\bf 57.84} && SCN & 99.72 & \bf 99.71 & 96.52 & 88.49 & {\bf 76.57} \\ [-0.45ex]
CapsNet  & 99.34 & 90.58 & 72.39 & 56.63 & 44.81 & & CapsNet & 99.60 & 99.37 & 94.77 & 79.53 & 60.22  \\ [-0.45ex]
SR-Capsule  & 99.26 & 91.53 & 76.26 & 56.74 & 49.39 & & SR-Capsule & 99.41 & 99.38 & 98.10 & 82.91 & 63.78\\[-0.45ex]
IDPACapsule & 99.56 & 94.71 & 77.45 & 60.43 &  51.81 & & IDPACapsule &  99.63 & 99.52 & 97.30 & 85.56 & 66.14  \\[-0.45ex] \rowcolor{LightCyan} 
HGWCapsule &  \bf{99.73} &  \bf{97.21} & \bf{80.56} & \bf{65.75} & 55.32 & \cellcolor{white}& HGWCapsule & \bf{99.87} &  99.70 & \bf 98.56 & \bf{92.34} & 71.32\\ [-0.45ex]
\cline{1-6}\noalign{\vskip\doublerulesep
         \vskip-\arrayrulewidth} \cline{1-6}\cline{8-13} \noalign{\vskip\doublerulesep
         \vskip-\arrayrulewidth} \cline{8-13}
\multirow{2}{*}{Method} &\multicolumn{5}{c}{\textbf{Training on Transformed Data} (2, $60^\circ$)} & &\multirow{2}{*}{Method} &\multicolumn{5}{c}{\textbf{Training on Transformed Data} (2, $90^\circ$)}\\ 
\cline{2-6}\cline{9-13}
   & \hfil (0, $0^\circ$) & \hfil (2, $30^\circ$) & \hfil (2, $60^\circ$) & \hfil (2, $90^\circ$) &  (2, $180^\circ$) & & 
   & \hfil (0, $0^\circ$) & \hfil (2, $30^\circ$) & \hfil (2, $60^\circ$) & \hfil  (2, $90^\circ$) & (2, $180^\circ$) \\  \cline{1-6}\cline{8-13}
SCN  & 99.64 & 99.41 & 98.72 & \bf98.49 & 76.50 & & SCN & 99.38 & 99.21 & 98.85 & 96.63 &  84.76\\ [-0.45ex]
CapsNet   & 99.23 & 99.18 & 97.34 & 94.20 & 73.62 & & CapsNet & 99.17  & 98.65 &  98.61 & 98.33 & 79.56  \\ [-0.45ex]
SR-Capsule &  98.74 & 98.28 & 97.31 & 93.52 & 79.52 & & SR-Capsule & 98.83 & 97.59 & 97.25 & 96.88 &  80.42 \\[-0.45ex]
IDPACapsule  & 99.58 & \bf99.56 & 98.82 & 97.36 & 77.85 & & IDPACapsule & 99.34 &  99.28 & 99.03 & 98.54 & 79.68   \\ [-0.45ex] 
\rowcolor{LightCyan} 
HGWCapsule &  \bf{99.70} &  99.51 &  \bf99.45 &  97.86 &  \bf82.49 &\cellcolor{white} & HGWCapsule & \bf99.54 & \bf99.50 & \bf99.48 & \bf99.36 & \bf87.13 \\  [-0.45ex]
\cline{1-6}\noalign{\vskip\doublerulesep
         \vskip-\arrayrulewidth} \cline{1-6}\cline{8-13} 
\multirow{2}{*}{Method} & \multicolumn{5}{c}{\textbf{Training on Transformed Data} (2, $180^\circ$)}\\ \cline{2-6}
   & \hfil (0, $0^\circ$) & \hfil (2, $30^\circ$) & \hfil (2, $60^\circ$) & \hfil (2, $90^\circ$) &  (2, $180^\circ$)  \\ 
 \cline{1-6}
SCN & 98.17 & 97.66 & 96.73 & 97.28 & 97.12  \\ [-0.45ex]
CapsNet & 96.52 & 96.55 & 96.51 & 95.84 & 95.78  \\  [-0.45ex]
SR-Capsule & 97.39 & 97.22 & 97.06 & 96.81 & 96.43\\ [-0.45ex]
IDPACapsule & 98.46 &  98.53 & 98.22 & 97.78 & 97.69  \\  [-0.45ex]  \rowcolor{LightCyan} 
HGWCapsule  &  \bf98.68 & \bf98.54 & \bf98.41 & \bf98.28  & \bf98.03\\ [-0.45ex]
\cline{1-6}
\end{tabular}
\end{table*}
\begin{table*}
\centering
\caption{Affine transformation experiments on CIFAR-10 (values are in \%)}
\label{tab:2}
\begin{tabular}{c|ccccclc|ccccc} \cline{1-6}\cline{8-13} 
\multirow{2}{*}{Method}& \multicolumn{5}{c}{\textbf{Training on Untransformed Data}} & &\multirow{2}{*}{Method}& \multicolumn{5}{c}{\textbf{Training on Transformed Data} (2, $30^\circ$)} \\ \cline{2-6}\cline{9-13} 
  & \hfil (0, $0^\circ$) & \hfil (2, $30^\circ$) & \hfil (2, $60^\circ$) & \hfil (2, $90^\circ$) &  (2, $180^\circ$) & &    & \hfil (0, $0^\circ$) & \hfil (2, $30^\circ$) & \hfil (2, $60^\circ$) & \hfil (2, $90^\circ$) &  (2, $180^\circ$) \\ \cline{1-6}\cline{8-13}
SCN & 89.64 & 52.83 & 48.12 & 44.66 & 43.52 && SCN  & 88.31 &  87.35 & 81.59 & 73.52 & 67.64 \\[-0.45ex]
CapsNet & 71.75 & 56.62 & 43.95 & 39.81 & 34.74 &&  CapsNet  & 72.55 & 69.17 & 62.29 & 52.87 & 46.67  \\ [-0.45ex]
SR-Capsule  & 87.54 &  63.25 & 47.78 & 49.67 & 45.26 & & SR-Capsule  & 86.35 & 83.74 & 78.51 & 74.30 & 68.92  \\ [-0.45ex]
IDPACapsule & 90.81 &  68.16 & {\bf 56.81} & 51.59 & {\bf51.37} && IDPACapsule & 90.29 & {\bf 89.78} & {\bf 88.65} & 76.51 & 70.32 \\ [-0.45ex] \rowcolor{LightCyan} 
HGWCapsule  & \bf92.54 & \bf73.58 & 53.43 & \bf51.76 & 48.56 & \cellcolor{white} & HGWCapsule &  \bf91.23 & 88.36 & 86.17 & \bf79.52 & \bf73.24	 \\ [-0.45ex]
\cline{1-6}\noalign{\vskip\doublerulesep
         \vskip-\arrayrulewidth} \cline{1-6}\cline{8-13} \noalign{\vskip\doublerulesep
         \vskip-\arrayrulewidth} \cline{8-13}
\multirow{2}{*}{Method} &\multicolumn{5}{c}{\textbf{Training on Transformed Data }(2, $60^\circ$)} & &\multirow{2}{*}{Method} &\multicolumn{5}{c}{\textbf{Training on Transformed Data }(2, $90^\circ$)} \\ 
\cline{2-6}\cline{9-13}
   & \hfil (0, $0^\circ$) & \hfil (2, $30^\circ$) & \hfil (2, $60^\circ$) & \hfil (2, $90^\circ$) &  (2, $180^\circ$) & &    & \hfil (0, $0^\circ$) & \hfil (2, $30^\circ$) & \hfil (2, $60^\circ$) & \hfil (2, $90^\circ$) &  (2, $180^\circ$) \\ \cline{1-6}\cline{8-13}
SCN  & 82.96 & 83.58 & 83.16 & 81.73 & 76.51 && SCN & 83.37 & 82.55 & 82.21 & 81.40 & 80.35  \\ [-0.45ex]
CapsNet  & 70.56 & 67.31 & 66.61 & 63.42 & 51.23 && CapsNet & 67.24 & 65.51 &  65.37 & 64.58 & 57.86   \\ [-0.45ex]
SR-Capsule  & 85.44 & 85.21 & 84.28 & 81.14 & 78.63  && SR-Capsule & 86.62 & { 85.41} & 82.66 & 81.62 & 79.23    \\ [-0.45ex]
IDPACapsule  & 86.91 & 87.27 & {\bf 86.51} & 82.26 & 80.45 && IDPACapsule  &  85.30 & 83.68 & 83.55 & 83.19 & {81.26} \\ [-0.45ex] \rowcolor{LightCyan} 
HGWCapsule  & \bf89.38 & \bf88.22 &  86.27 & \bf84.23 &  \bf82.78 & \cellcolor{white}& HGWCapsule & \bf87.24 & \bf 85.93 & \bf84.77 & \bf82.91 & {\bf 81.38}	\\ [-0.45ex]
\cline{1-6}\noalign{\vskip\doublerulesep
         \vskip-\arrayrulewidth} \cline{1-6}\cline{8-13} 
\multirow{2}{*}{Method} & \multicolumn{5}{c}{\textbf{Training on Transformed Data} (2, $180^\circ$)}\\ \cline{2-6}
  & \hfil (0, $0^\circ$) & \hfil (2, $30^\circ$) & \hfil (2, $60^\circ$) & \hfil (2, $90^\circ$) &  (2, $180^\circ$) \\  \cline{1-6}
SCN & 81.50 & 80.58 & 80.64 & 80.73 & 80.21 \\[-0.45ex]
CapsNet  & 64.76 & 61.59 & 61.04 & 59.86 & 59.97   \\ [-0.45ex]
SR-Capsule  & 82.20 & 80.79 & 81.06 & 80.46 & 80.32 \\ [-0.45ex]
IDPACapsule  & 84.23 & 83.55 & 82.41 & 82.16 & 81.39   \\ [-0.45ex] \rowcolor{LightCyan} 
HGWCapsule  & \bf84.56 & \bf83.71 & \bf83.52 & \bf83.95 & \bf83.80\\   [-0.45ex]
\cline{1-6}
\end{tabular}
\end{table*}

\section{Experiments}
\label{expriment}
In this section, we evaluate our model on images with different scales, backgrounds, and textures. 
We performed four sets of experiments: {\bf (Q1)} to show the robustness of HGWCapsule, we compare its performance with four strong CapsNet baselines on affine transformations using, MNIST \cite{lecun1998gradient}, CIFAR-10 \cite{krizhevsky2009learning}, and  SmallNORB \cite{lecun2004learning} datasets; {\bf (Q2)} to demonstrate the lightweight and applicability of our capsule, we combine it with SOTA CNN backbones for different tasks,  including classification on ImageNet \cite{deng2009imagenet}, and CIFAR-100 \cite{krizhevsky2009learning}, semantic segmentation on Pascal voc \cite{everingham2015pascal}, ISIC-2018 \cite{codella2019skin}, object detection on COCO \cite{chen2015microsoft}, and image matching on Aachen  \cite{sattler2018benchmarking}; {\bf (Q3)} we also show the computation cost of the proposed model against several capsule baselines; {\bf (Q4)}  we conducted an ablation study to evaluate the impact of different factors on the overall performance of our model.

\begin{figure*}
\includegraphics[height=6.2cm]{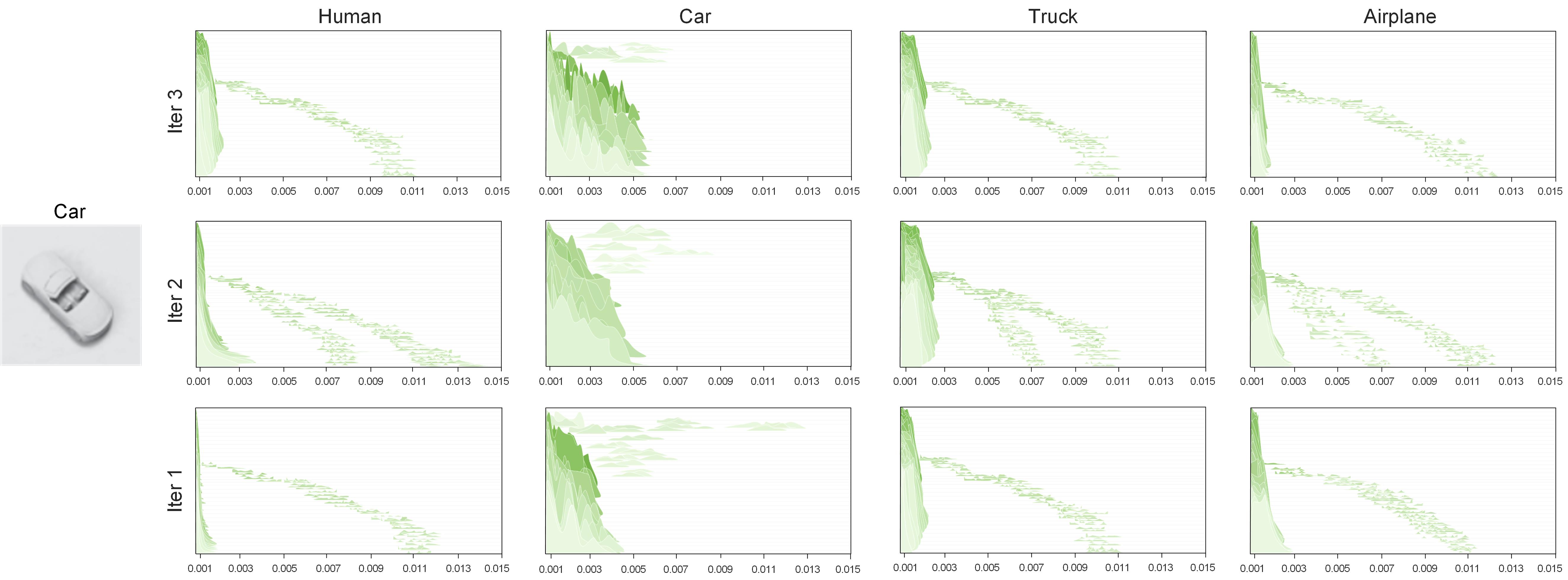} 
\caption{Histograms of the distances ({\bf X} axis) between the {\bf input} (car) and of each of the 4 class capsules (airplane, car, truck, human) throughout training (epochs on {\bf Y} axis) on the smallNORB dataset. The histograms visualize the frequency distributions of the input to the embedding of each capsule during training. Iterations 1–3 are depicted per row, and each column represents a different class capsule. In each row, the HGW mechanism learns to detect the best alignment for the input sample, i.e., the correspondence between input and class capsules. As can be observed, the car image (input) correctly routes with the target capsule during training, and the discrepancies between their distributions gradually decrease over iterations. }
\label{fig:20}
\end{figure*}

\subsection{Robustness to Affine Transformations (Q1)}
\subsubsection{Experiments on MNIST and CIFAR-10. }
\begin{figure}
\centering
\includegraphics[height=3.1cm]{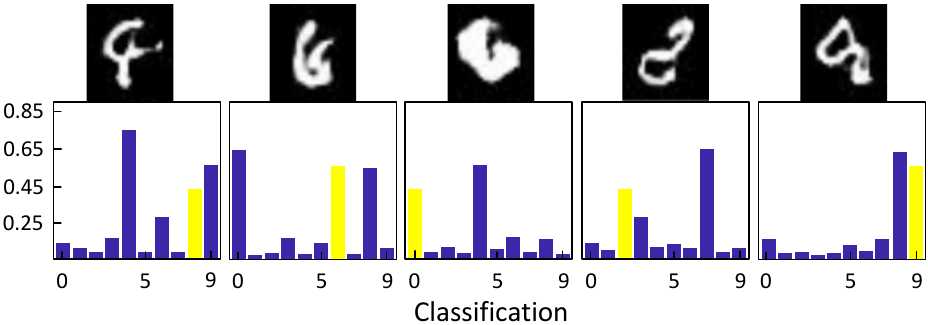} 
\caption{Digits misclassified using HGWCapsule. Correct classes are indicated by yellow bars whose height is the length of the corresponding capsule. These samples are ambiguous, and hence it is assumed that the error cannot be 0 but is around (0.10 - 0.15)\%.} 
\label{fig:mis}
\end{figure}

This section aims to evaluate the capsule baselines' robustness on affine transformations involving rotation and translation on MNIST and CIFAR-10. We created five different versions of training and testing sets by randomly transforming them, such that we have $[0 \text{ pixel}, 0^\circ]$, $[2\text{ pixel}, 30^\circ]$, $[2\text{ pixel}, 60^\circ]$, $[2\text{ pixel}, 90^\circ]$, and $[2\text{ pixel}, 180^\circ]$. We train each model using its training sets and evaluate their performance on the respective testing sets. As a result, each model obtained 25 accuracy results on each dataset. We compare our model, which has 0.53M parameters, with four strong Capsule-based methods, including CapsNet (8.5M) \cite{sabour2017dynamic}, SR-capsule (4.1M) \cite{pucci2021self}, IDPACapsule (1.5M) \cite{tsai2020capsules}, and SCN (3.8M) \cite{edraki2020subspace}. The results are shown in Tables \ref{tab:1} and \ref{tab:2}. 
Results of this experiment suggest that our proposed HGWCapsule is obviously more robust to input geometric changes during training and testing. The HGWCapsule can handle the extreme transformations of the training data and generalizes better to the testing set in most cases. Some baselines, however, outperform our model in certain splits.
On CIFAR-10, IDPACapsule slightly outperforms our model in 5 out of 25 experiments. IDPACapsule, which is based on the self-attention concept, performs better than HGWCapsule on the untransformed set or with minor training transformations. HGWCapsule, on the other hand, excels when applied to extreme training transformations (e.g., 90 or 180-degree rotations), outperforming IDPACapsule in most other cases. Our observation is that since the attention mechanism can mainly capture long-range relationships between features, small shifts in the positions of the features can disrupt the learned relationships and result in degraded performance in certain cases. IDPACapsule's parameters are 3$\times$ greater than HGWCapsule. 
Regarding MNIST, SCN outperforms HGWCapsule in 4 out of 25 cases. Specifically, SCN exhibits better generalization to test-set transformations compared to HGWCapsule. However, in most other cases, our model surpasses SCN in learning from extreme training perturbations. The first possible reason for this performance is the greater number of parameters in the SCN (almost seven times higher than ours), which allows the model to learn features that generalize more effectively across various transformations. Further, the simple prediction mechanisms used by SCN may be a factor in its inability to perform well under extreme training perturbations. The main focus of SCN is to reduce computational complexity by projecting data onto a lower-dimensional space rather than capturing all the underlying information of the data. Moreover, the digit dataset contains ambiguous images that are difficult to classify correctly. An example of these ambiguous digits is shown in Fig. \ref{fig:mis}, where our model struggles to capture the representative labels effectively. To ensure the length of the capsule is bound in [0, 1], we use $c_{ij}=\frac{1}{w_{i j}+1}~ w_{i j}$. 

\begin{figure}
\centering
\includegraphics[height=5.5cm]{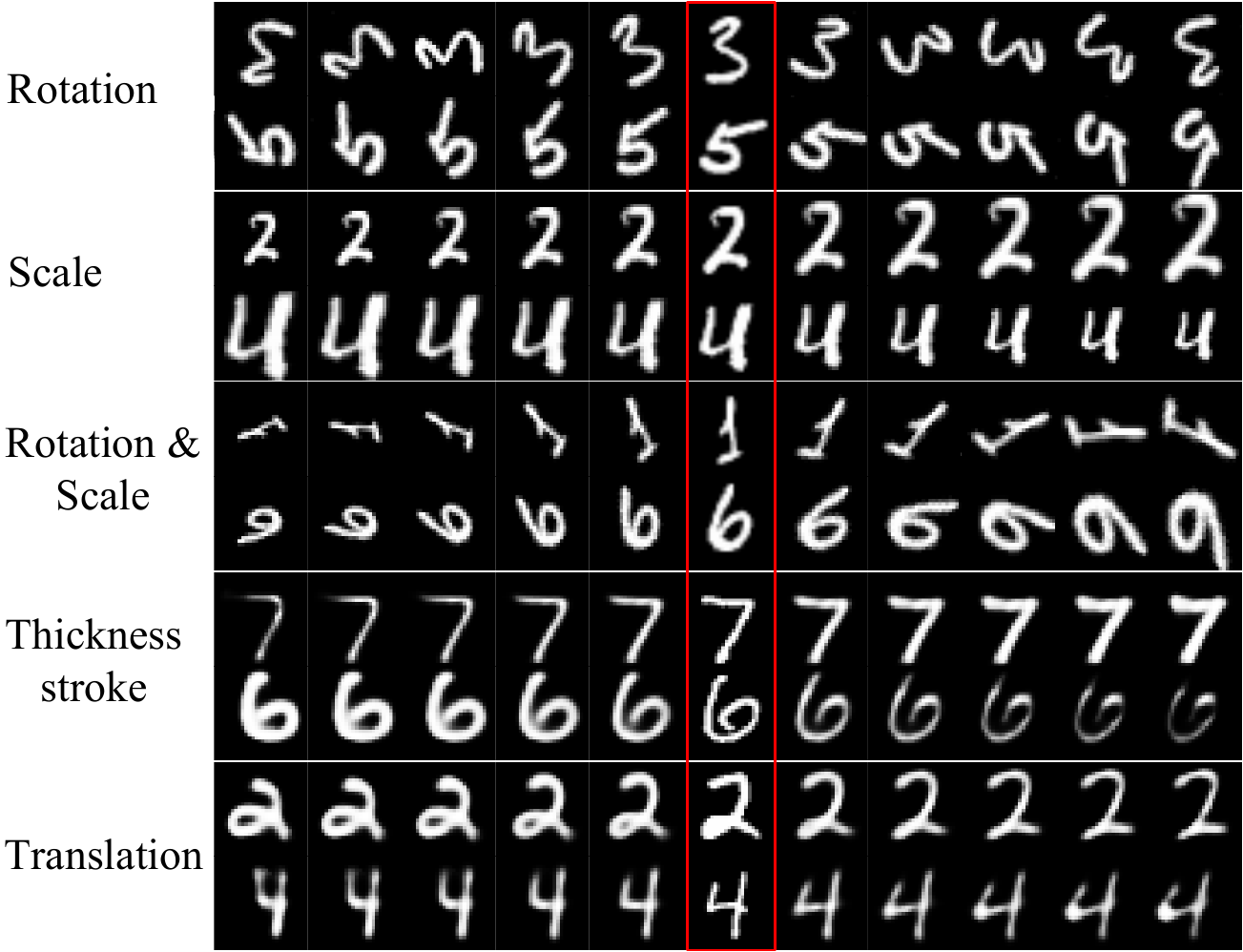}
\caption{Each row shows one features like rotation, scaling, thickness, and translation of the digits achieved by tweaking capsule dimensions in the first layer within the range of  [-2.5, 2.5]. The red column indicates the input. The samples in each row are diverse, and by adjusting the capsule dimension, we can change the appearance manifold of each digit. } \label{f:trans}
\end{figure}

In addition to the above experiments, we also demonstrate the ability of our model to produce different affine transformations. Fig.~\ref{f:trans} showcases this property of our model. Each row represents one feature like rotation, thickness, and scale of the digits, and samples are generated by tweaking one dimension of capsules of the first generator layer in the range of [-2.5, 2.5].\\ 

\begin{figure*}
\centering
\includegraphics[height=8.5cm]{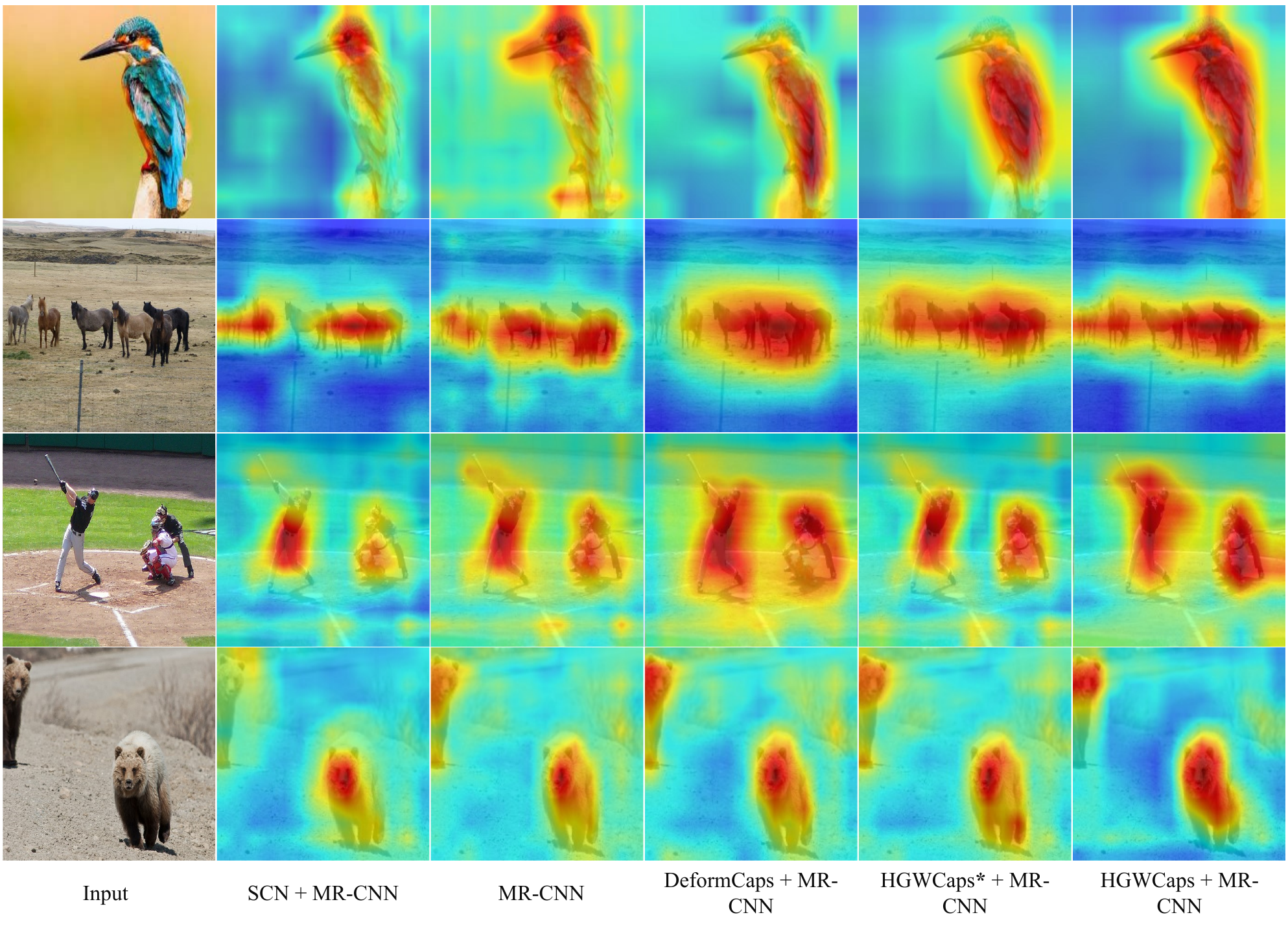}
\caption{Visualization of different CapsNet and CNN approaches. The results were obtained for four random samples from the MS COCO test-dev dataset and were compared for a baseline MRCNN \cite{he2017mask}, SCN \cite{edraki2020subspace}, DeformCaps \cite{lalonde2021deformable} and two variants of our model. HGWCaps* denotes our proposed model trained without using regularization term in Eq.\ref{eq:21}. To fairly compare CNN and CapsNets, we use the same backbone, ResNet-50 for all five methods. } \label{attention}
\end{figure*}

\begin{figure}
\centering
\includegraphics[height=9.2cm]{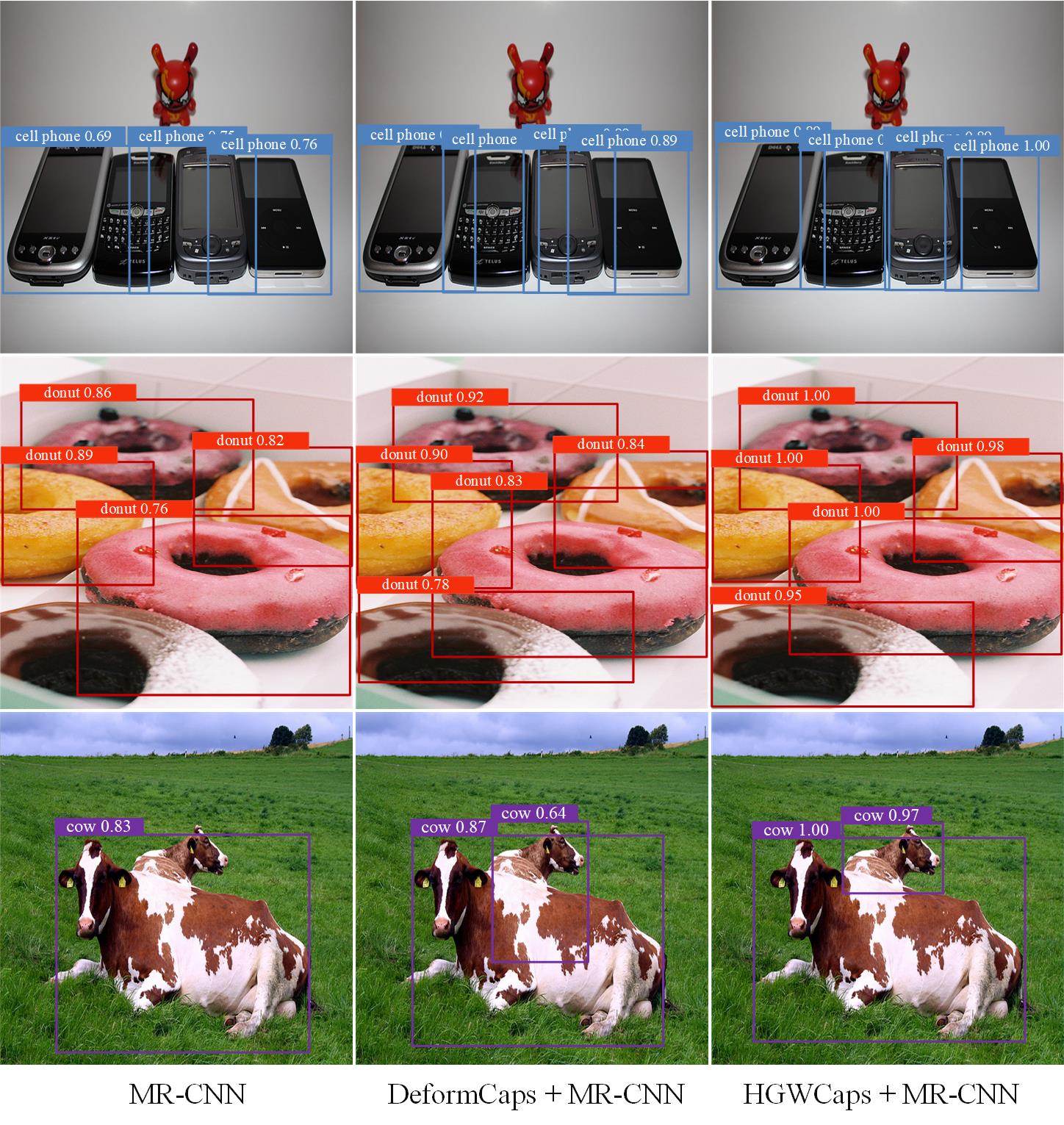}
\caption{Qualitative results for MRCNN\cite{he2017mask} (leftmost), DeformCapsule \cite{lalonde2021deformable} (centre), and Our proposed model (rightmost) on the MS COCO test-dev dataset. Our model achieves higher average precision (AP) values than DeformCaps and MRCNN. However, DeformCapsule obtains slightly higher AP values than MRCNN, but it also generates more false positives. To fairly compare CNN and CapsNets, we use the same backbone, ResNet-50 for all these methods. } \label{obj1}
\end{figure}

\subsubsection{Experiments on SmallNORB}
SmallNORB is a dataset for 3D object recognition from shapes with 5 toy classes, such as cars, airplanes, trucks, humans, and animals. We selected smallNORB because it is much closer to natural images than MNIST and is a pure shape recognition task unaffected by context or color. For this experiment, we follow the setting of \cite{hinton2018matrix} by resizing the training images to 32$\times$32 pixels without applying brightness augmentation during training. Since it is less complicated than CIFAR-100, we use a shallower network that consists of two convolution layers, which are then followed by three successive capsule layers, trained for 250 epochs with Adam optimizer and learning rate of 3e-3 with exponential decay, and achieve {\bf 1.17\%} test error. The best results on smallNORB are reported for the ``CapsNet" \cite{sabour2017dynamic} {\bf 3.77\%}, {\bf 1.82\%} for the ``EMCapsule" \cite{hinton2018matrix}, {\bf 2.26\%} and  {\bf 1.63\%} for the ``E-CapsNet" \cite{mazzia2021efficient} and ``SR-Capsule" \cite{hahn2019self}, respectively. 
In Fig.~\ref{fig:20}, we show how our model adjusts the interactions between the input and a group of capsules to achieve their optimal correspondence. The histograms visualize the frequency distributions of the input to the embedding of each capsule during training. Results are visualized for only alignments with distances less than 0.03, and the three rows show iterations 1, 2, and 3. The leftmost column is our input image (car), while the other four represent different class capsules. Our routing procedure learns to capture the best alignment between the input and class capsules in each row. As can be seen, the car image correctly routes with the target capsule in the second and third iterations. This means that over iterations, the discrepancies between the input and the target capsule are increasingly approaching 0 compared to the other class capsules. However, the first iteration shows a failure case, where the car is confused with a truck. 

\subsection{Comparison with CNN-based Networks on Advanced Vision Tasks (Q2)}
We conduct extensive experiments on various benchmarks:  CIFAR-10/100, ImageNet-1K for image classification, MS COCO, PASCAL VOC, and ISIC2018 for object detection and segmentation, and Aachen for image matching. For fairness, we used PyTorch to retrain all the models and provide the reproduced results. \\

\subsubsection{Classification Results on CIFAR-10/100}
\label{sec1}
In Tables \ref{tab:c1}, and \ref{tab:c2}, we compare the performance on CIFAR-10/100  \cite{krizhevsky2009learning}  after placing capsule layers at the bottlenecks of state-of-the-art models including ResNet-110 \cite{he2016deep}, WideResNets-28 \cite{zagoruyko2016wide}, and DenseNet-BC-100 \cite{jegou2017one}. The Wide-ResNet-28-10 architecture consists of three stages, each containing four residual blocks that use two $3\times3$ convolutions. We replace the second convolution of all residual blocks with the capsule layers while the batch normalization layers and residual connections of this block have been removed. Mean pooling is substituted with capsule pooling-- for capsules that have the same visual attribute, a single capsule with the mean of those capsule vectors can represent all of them. 
The training process utilized SGD for 100 epochs, with a momentum rate 0.9. The initial learning rate was set to 0.1 and decreased by 0.1 after every 30 epochs. The core building block of a ResNet is a bottleneck block, where each bottleneck block contains three convolution layers: a $1\times1$, a $3\times3$, and another $1\times1$. We also update the $3\times3$ convolution layers in the last four bottleneck blocks (i.e., the final stage) with capsule layers, each consisting of 32 capsule types with a dimension of 2. The batch normalization layers and residual connections of these blocks have been removed. Capsule pooling replaces mean pooling (as we discussed above). 
However, we performed two modifications when considering the other backbone (DenseNet). First, we replace the 3$\times$3 convolution used in the last dense blocks with the capsule layers, and bath normalization layers in these blocks are removed. Then, the mean pooling is replaced with our capsule pooling in the last two transition layers. While ResNet and WideResNet achieve 78.44\% and 81.29\% top-1 accuracy, respectively, these backbones with capsule layers achieve an accuracy of 79.82\% and 82.93\% using a lower number of the parameters. The efficiency and effectiveness of our capsule in enhancing network capacity while maintaining relatively fewer parameters are attributed to two main factors.
i) Our proposed framework allows the capsule to route information to the most relevant capsules in the next layer while suppressing irrelevant ones. This ensures computational resources are focused on relevant information, reducing redundancy in network parameters. ii) CapsNets can also reduce the multiply-add operation by using fewer filters in the capsule layer than standard convolution layers. Each capsule layer represents multiple object attributes, such as position, orientation, and scale, leading to a more compact representation of a wide range of object attributes than convolution filters. \\

\begin{table}
\centering
\caption{Results of image classification on CIFAR-10} 
\label{tab:c1}
\begin{tabular}{c|ccc}    \hline\hline
\rule{0pt}{0.8\normalbaselineskip}Method  &top-1 acc. & MAC. & \# params \\  \hline
\rule{0pt}{0.8\normalbaselineskip}ResNet  & 93.57 & 255.23M & 1.72M \\  \rowcolor{LightCyan} 
HGWCapsule + ResNet    & {95.18} & {225.28M} & {1.19M}  \\[0.2ex]\hline  \rowcolor{Gray}
 Relative improvement &  {\bf +1.61\%}  & \bf 14\% & \bf 30\% \\ \hline 
\rule{0pt}{0.8\normalbaselineskip}DenseNet &95.32  & 296.38M & 769.75K   \\   \rowcolor{LightCyan} 
HGWCapsule + DenseNet &{95.78} & {268.24M} & {711.21K}   \\ [0.2ex]\hline  \rowcolor{Gray}
 Relative improvement &  {\bf +0.46\%} & \bf 9\% &  \bf 8\% \\ \hline 
\rule{0pt}{0.8\normalbaselineskip}WiderResNet  & 95.89 & 5.25G & 36.45M  \\  \rowcolor{LightCyan} 
HGWCapsule + WiderResNet  &{96.37} & {3.81G} & { 22.19M} \\  \hline   \rowcolor{Gray}
 Relative improvement &  {\bf +0.48\%} & \bf 27\% &   \bf 38\% \\ \hline 
\end{tabular}
\end{table}
\begin{table}
\centering
\caption{Results of image classification on CIFAR-100} 
\label{tab:c2}
\begin{tabular}{c|ccc}    \hline\hline
\rule{0pt}{0.5\normalbaselineskip}Method & top-1 acc. & MAC. & \# params \\ [0.4ex]\hline
\rule{0pt}{0.8\normalbaselineskip}ResNet & 78.54 & 255.24M & 1.75M \\  \rowcolor{LightCyan} 
HGWCapsule + ResNet & {79.82} & {226.86M} & {1.22M}  \\[0.2ex]\hline \rowcolor{Gray}
 Relative improvement &  {\bf +1.28\%} & \bf 14\%  & \bf 31\%  \\ \hline  
\rule{0pt}{0.8\normalbaselineskip}DenseNet & 77.23  & 286.58M & 800.28K   \\   \rowcolor{LightCyan} 
HGWCapsule + DenseNet & {78.16} & {267.43M} & {716.45K}   \\[0.2ex] \hline  \rowcolor{Gray}
 Relative improvement &   {\bf +0.93\%}  & \bf 8\% & \bf11\% \\ \hline 
\rule{0pt}{0.8\normalbaselineskip}WiderResNet & 81.29 & 5.25G & 36.72M  \\  \rowcolor{LightCyan} 
HGWCapsule + WiderResNet & {82.93} & {3.90G} & {23.83M} \\  \hline  \rowcolor{Gray}
 Relative improvement &   \bf +1.64\% & \bf 25\% & \bf 35\% \\ \hline 
\end{tabular}
\end{table}

\setlength{\tabcolsep}{15pt}
\begin{table}
\centering
\caption{The single crop top-1/5 error rate results on ImageNet-1k, with different CapsNet and CNN baselines. ResNet \cite{he2016deep}, SqueezeNet \cite{hu2018squeeze}, and AA-ResNet  \cite{bello2019attention} results are obtained from the their GitHub page.  } 
\label{tab:imagenet}
\begin{tabular}{l|ccc}  \hline\hline
\rule{0pt}{1\normalbaselineskip}Methods &  Backbone &top-1 & top-5  \\  \hline 
\rule{0pt}{0.8\normalbaselineskip}ResNet-101 \cite{he2016deep} & &21.89  & 6.22\\
ResNet-50 \cite{he2016deep} & & 23.58 & 6.73 \\
SE-ResNet-50 \cite{hu2018squeeze}& &22.81 & 6.52 \\
AA-ResNet-50 \cite{bello2019attention}& &22.56 & 6.28 \\ [0.5ex] \hline
\rule{0pt}{0.8\normalbaselineskip}SCN \cite{edraki2020subspace}& ResNet-50 & 24.79 & 7.35\\
Trans-Caps \cite{mobiny2019automated}& ResNet-50 & 29.47 & 9.78 \\
STAR-Caps \cite{ahmed2019star}& ResNet-50 & 23.57 & 7.04\\
IDPA-Caps\cite{tsai2020capsules}& ResNet-50& 23.48 & 6.81 \\ \rowcolor{LightCyan} 
HGWCaps(ours) & ResNet-50 & \bf21.85 &  \bf6.21  \\ \hline  
\end{tabular}
\end{table}

\subsubsection{Classification Results on ImageNet}
\label{sec2}
ImageNet \cite{deng2009imagenet} is one of the largest and most complex image classification benchmarks, and we show the effectiveness of our capsule in such a difficult task. We
use the ResNet-50 architecture \cite{he2016deep} because of its widespread use and its ability to easily scale across several computational budgets. ResNet-50 uses a bottleneck block comprising of 1$\times$1, 3$\times$3, 1$\times$1 convolutions where the last convolution expands the number of filters and the first one contracts the number of filters. We modify the ResNet by replacing the 3$\times$3 convolutions with our capsule, as this decreases the number of parameters. We apply HGWCapsule in each residual block of the last two stages of the architecture. Table \ref{tab:imagenet} compares our model's performance with several CNN and CapsNet baselines.
HGWCapsule significantly outperforms the original backbone and even surpasses the Squeeze-and-Excitation (SE) \cite{hu2018squeeze} while using fewer parameters (27.6M$\rightarrow$19.2M). Interestingly, our HGWCaps-ResNet-50 performs comparably to the baseline ResNet-101, suggesting that the performance improvement is not solely due to increased depth by adding more layers to bottlenecks. Incorporating the scalable capsule layers into deep architectures provides us with both the advantages of a low number of parameters and high performance. Other CapsNets such as DRCapsNet \cite{sabour2017dynamic}, CVAE-Capsule \cite{guo2021conditional}, and SR-Capsule \cite{hahn2019self} significantly underperform as dataset complexity increases and fail to converge on ImageNet. Therefore, we exclude them from this experiment. IDPA-Capsule and STAR-Capsule slightly outperform the ResNet backbone, but they couldn't perform as well as SE-ResNet and AA-ResNet. 

\setlength{\tabcolsep}{2.5pt}
\begin{table}
\centering
\caption{Object detection on COCO with different CNN and CapsNet baselines} 
\label{obj}
\begin{tabular}{c|cccccccc}  \hline\hline
& \multicolumn{6}{c}{Mask R-CNN (MRCNN)} \\  \hline 
\rule{0pt}{1\normalbaselineskip}Methods & \# param. & $\text{AP}^b$ & $\text{AP}_{50}^b$ & $\text{AP}_{75}^b$ & $\text{AP}^m$ & $\text{AP}_{50}^m$ & $\text{AP}_{75}^m$  \\  \hline 
\rule{0pt}{1\normalbaselineskip} ResNet-101 \cite{he2016deep} & 44.1M & 37.59 &  56.80 &  41.38 &  35.21 & 56.71 & 36.23  \\ \hline
\rule{0pt}{1\normalbaselineskip} DeformCaps \cite{lalonde2021deformable} & 38.2M & 43.36 & \bf65.74 & 51.29 & 41.83 & \bf 63.71 & 43.65\\ 
\rule{0pt}{0.8\normalbaselineskip} SCN \cite{edraki2020subspace} & 41.5M & 35.27 & 52.63 & 39.25 & 33.15 & 51.35 & 34.98 \\\rowcolor{LightCyan} 
HGWCaps  & \bf 35.7M & \bf 45.71 & 65.32 & \bf 52.73 & \bf 42.51 & 62.30 & \bf45.88 \\\hline
\end{tabular}
\end{table}
\begin{table}
\centering
\caption{Results of semantic segmentation on ISIC-2018 and PASCAL VOC 2012.} 
\label{tab:c3}
\resizebox{8.9cm}{!}{
\begin{tabular}{c|cc|cc|cc}    \hline\hline
\multirow{2}{*}{Method} & \multicolumn{2}{c|}{mIoU-ISIC}& \multicolumn{2}{c|}{mIoU-PASCAL}& \multicolumn{2}{c}{\# params} \\ [0.3ex]\cline{2-7}
\rule{0pt}{0.8\normalbaselineskip} & original & \cellcolor{LightCyan}{HGWCapsule} & original & \cellcolor{LightCyan}{HGWCapsule} & original & HGWCapsule \\[0.4ex] \hline
\rule{0pt}{0.5\normalbaselineskip}UNet & 67.23 &  \cellcolor{LightCyan}{\bf68.71} &78.20 &  \cellcolor{LightCyan}{\bf80.47}& 31.4M & {\bf23.7M}\\  
DeepLabv3+ & 81.58 & \cellcolor{LightCyan}{\bf82.33} &87.82 & \cellcolor{LightCyan}{\bf89.54} & 59.5M  & {\bf48.3M} \\
 \hline 
\end{tabular}
}
\end{table}

\subsubsection{Object Detection on COCO}\label{sec3}
We conduct object detection on the MS COCO dataset \cite{chen2015microsoft} using Mask R-CNN (MRCNN) \cite{he2017mask} as our detection method with ResNet-101 pretrained on ImageNet-1K as a backbone network. All models are trained on the 118k training images and evaluated on 5K validation images. We compare our model with other SOTA backbones, including MRCNN, SCN, DeformCapsule, and HGWCaps{\bf*}, which indicate our model without adding the regularizer term in Eq.~\ref{eq:21}. Here, we aim to improve detection performance by incorporating HGWCapsule into the baseline. Because we use the same detection method for all models, the gains can only be attributed to our HGWCaps+ResNet.
For all the compared models, the backbones were first pretrained using ImageNet-1K. The pretrained models are then used to fine-tune on the detection task and report the box/mask mAP ($AP^b$, $AP^m$) of different CNN and Capsule backbones. We followed the training procedure in \cite{lalonde2021deformable}, training on 512$\times$512 pixel inputs, yielding 128$\times$128 detection grids, using random flip, random scaling (between 0.6 and 1.3). We also use the AdamW \cite{kingma2014adam} optimizer with a learning rate of 5e-4 for 40 epochs (batch size of 12), with 5$\times$ drops at 5, 15, and 25 epochs. Our results are summarized in Table \ref{obj}, Fig. \ref{attention}, and Fig. \ref{obj1}. DeformCaps performs slightly better at the $\text{AP}_{50}$, and captures more objects, but also leads to a higher number of false positives (see Fig. \ref{obj1}). It means that some of the detections made by DeformCaps are less accurate or more likely to be incorrect. HGWCapsule's joint exploitation of feature and structural information enables it to focus on the precise localization of objects, leading to better results in cases where objects have partial or incomplete coverage ($\text{AP}_{25}$) or high overlap with the ground truth ($\text{AP}_{75}$). However, CNN's baseline obviously works better than SCN.

\begin{figure}
\centering
\includegraphics[height=5.5cm]{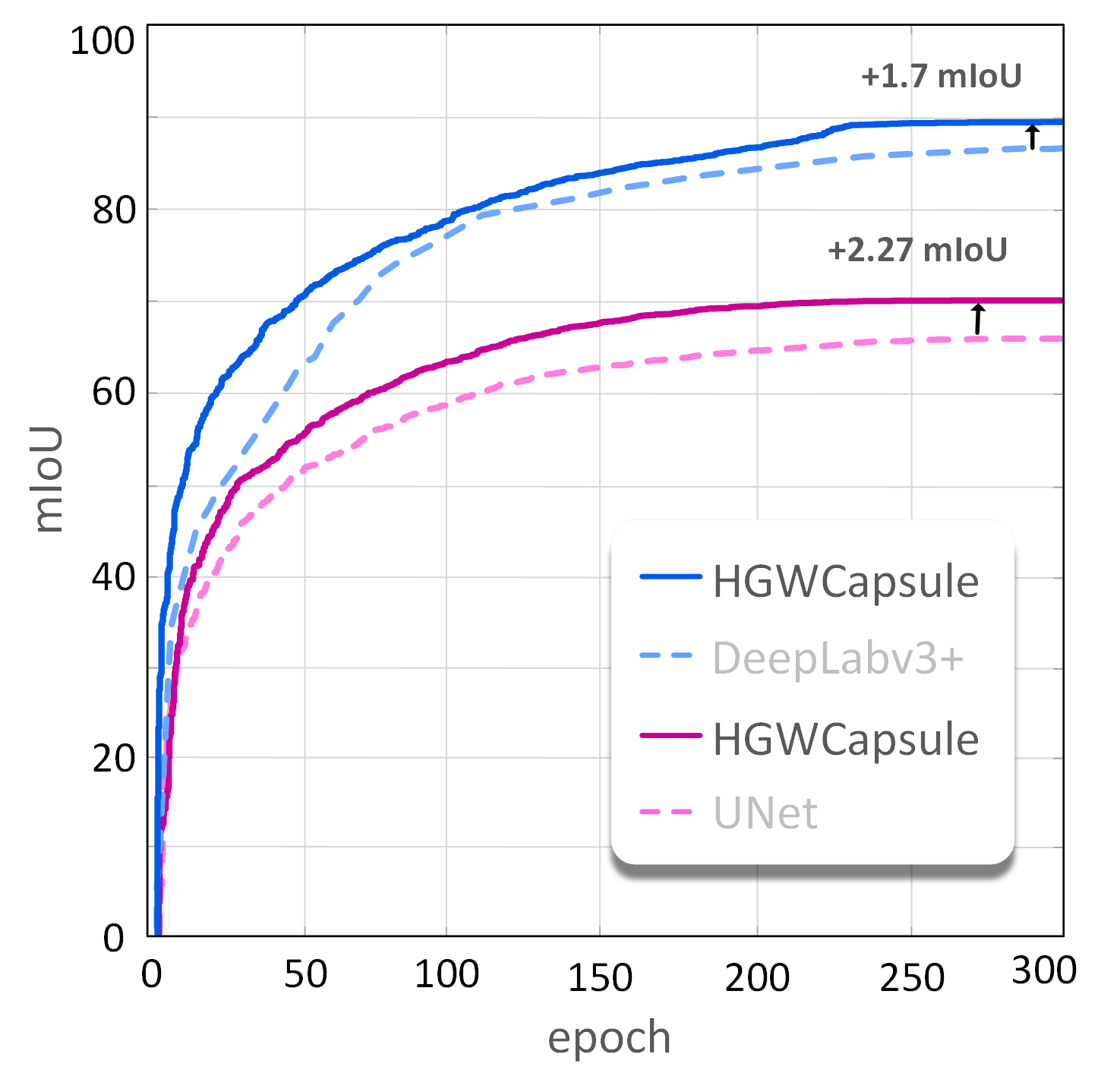} %
\caption{mIoU scores for HGWCaspule with UNet and DeepLabv3 backbones on the Pascal voc dataset. Our model outperforms the backbones while using fewer parameters.}  \label{pascal}
\end{figure}

\begin{figure*}
\centering
\includegraphics[height=8.7cm]{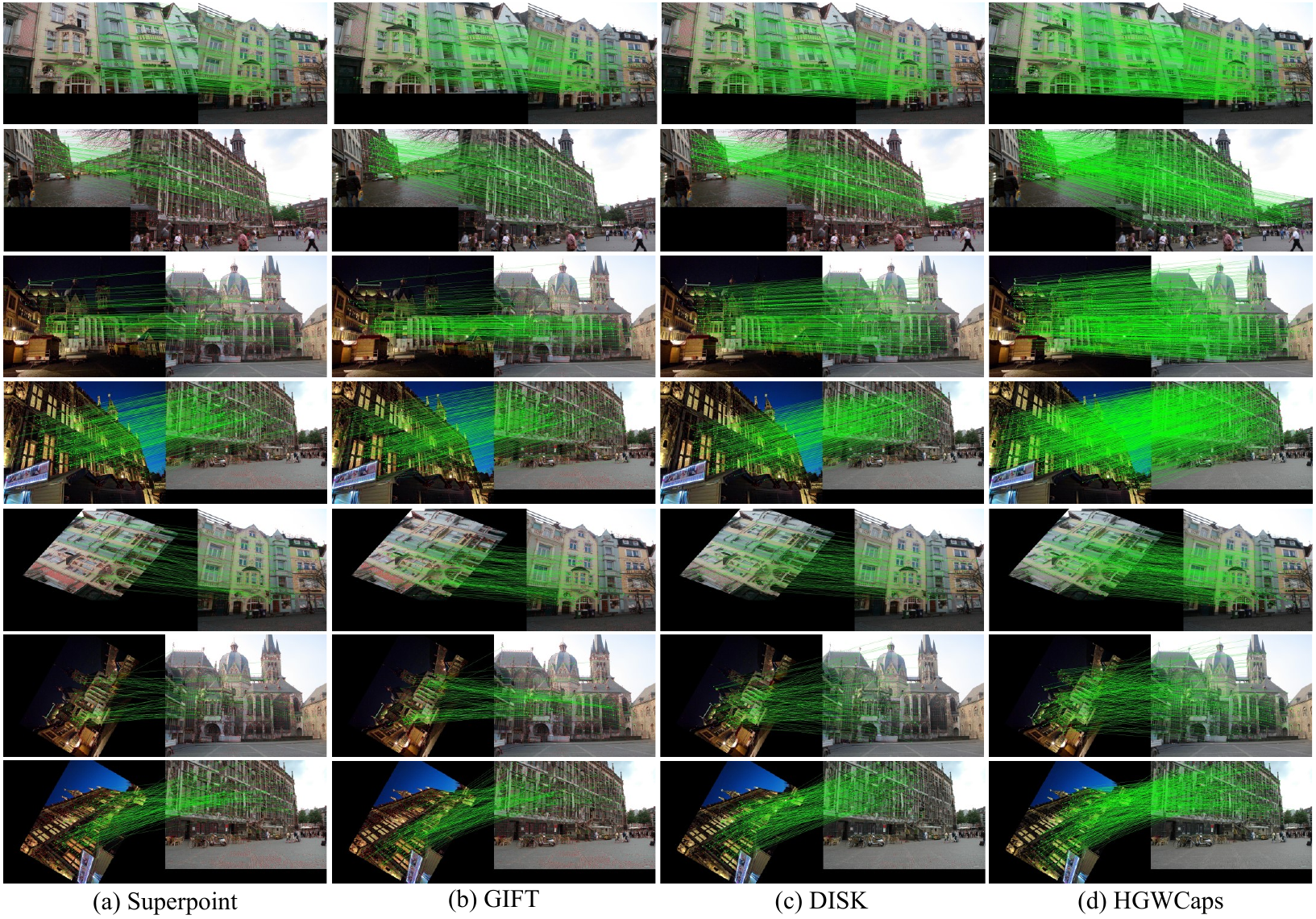} %
\caption{Visualization of estimated correspondences on Aachen \cite{sattler2018benchmarking}. The first column uses key points detected by Superpoint \cite{detone2018superpoint}, the second column uses GIFT \cite{liu2019gift}, the third column uses DISK \cite{tyszkiewicz2020disk}, and the last one is our proposed model. Our model obviously works better than baselines under extreme scale and orientation changes. }   
\label{fig:match}
\end{figure*} 

\subsubsection{Segmentation results}\label{seg}
The ISIC-2018 \cite{codella2019skin} and Pascal-VOC 2012 \cite{everingham2015pascal} datasets are used for semantic segmentation, and the baselines for this experiment are UNet \cite{ronneberger2015u} and DeepLabv3+ \cite{chen2017rethinking}. Table \ref{tab:c3} summarizes the segmentation results, including the number of parameters for each model and the mIoU (mean Intersection over Union). For HGWCaps-DeepLabv3, we further employ the ResNet-50 backbone, which has been pre-trained on ImageNet \cite{deng2009imagenet}. On the ISIC-2018, our model improves the UNet and DeepLabv3+ by +1.48\% and +0.75\%, respectively. However, the performance differs more on the challenging PASCAL VOC dataset, where our model outperforms the UNet and DeepLabv3+ by +2.27 and +1.72 while decreasing the model parameters by 25\% (31.4M $\rightarrow$ 23.7M) and 19\% (59.5M $\rightarrow$ 48.3M), respectively. DeepLabv3 is a strong segmentation method with its use of dilated convolutions and atrous spatial pyramid pooling, and it mainly focuses on capturing global context. HGWCapsule leverages the power of optimal transport and distance-based alignment to capture both local features and spatial relationships effectively. This makes HGWCapsule more robust in handling complex data compared to DeepLabv3+. We show the effect of HGWCapsule compared to CNN baselines by visualizing their convergence on the Pascal-voc in Fig. \ref{pascal}.

\subsubsection{Image matching on Aachen}\label{auc}
We use HGWCapsule to quantify the correspondences between two sample images under extreme scale and orientation changes. For this experiment, We use the Aachen dataset \cite{sattler2018benchmarking}, and compare the performance of our model with three strong methods; DISK \cite{tyszkiewicz2020disk}, GIFT \cite{liu2019gift}, and Superpoint \cite{detone2018superpoint}. The results are obtained by replacing the convolution layer in the baselines with the capsule layer to build the capsule-based models. We add rotation and synthetic scaling to images in Aachen dataset \cite{sattler2018benchmarking} and report the matching performances under extreme scale and orientation changes in Fig. \ref{fig:match}. We also compare the HGWCapsule with three baselines: DISK \cite{tyszkiewicz2020disk}, GIFT \cite{liu2019gift}, and Superpoint \cite{detone2018superpoint}. GIFT used the Superpoint \cite{detone2018superpoint} technique for keypoint detection, an expensive model that generates several unused key points. In HGWCapsule, we use \cite{altwaijry2016learning} for keypoint generation, which only considers those key points that went through geometric filtering. We modify the network architecture of \cite{altwaijry2016learning} by replacing the convolution layer of the Keypoint Scoring module with an HGWCapsule and replacing the last two convolution layers of the Patch Matching module with two Capsule layers. Our 3-layer capsule trained on Aachen performs better than the baseline convolution models. Our proposed model meets or beats the baselines on different tasks, demonstrating that the performance gain and robustness are mainly due to the proposed learning strategy. 

\begin{table*}[h!]
\centering
\caption{Ablation study. We show the classification error rates on SVHN. We achieve the best accuracy relative to CapsNet and CNN baselines even in a limited data regime.} 
\label{tab:abl}
\begin{tabular}{c|c|ccccc}    \hline\hline
\multirow{2}{*}{Routing Method} & \multirow{2}{*}{backbone} & \multicolumn{4}{l}{~~~~ \# training samples}\\ \cline {3-6}
 & &10k & 30k & 60k & 600k  \\ \cline {1-6}
\rule{0pt}{0.8\normalbaselineskip}EM Routing \cite{hinton2018matrix}  & simple& 13.78 & 12.39 & 11.81 & 9.75 \\
Dynamic Routing \cite{sabour2017dynamic} & simple & 12.78 & 11.26 & 11.45 & 9.32 &  \\
Self Attention Routing \cite{hahn2019self} & simple & 11.84 & 10.31 & 9.86 & 8.42 \\ 
Inverted Dot-Product Attention Routing \cite{tsai2020capsules} & simple & 10.91 & 9.82 & 9.25 & 7.31 \\\rowcolor{LightCyan} 
Hybrid Gromov-Wasserstein Routing -A &simple& 11.36 & 11.88 & 9.62 & 8.28  \\\rowcolor{LightCyan} 
Hybrid Gromov-Wasserstein Routing -B & simple& 11.71 & 11.26 & 10.59 & 9.73 \\\rowcolor{LightCyan} 
Hybrid Gromov-Wasserstein Routing -C & simple& 9.81 & 9.73 & 9.12 & 7.25 \\\rowcolor{LightCyan} 
Hybrid Gromov-Wasserstein Routing (ours) &simple& {\bf 9.68} & {\bf 9.24} & {\bf 8.53} & {\bf 6.82} \\ [0.5ex] \cline {1-6}
\rule{0pt}{1\normalbaselineskip} EM Routing \cite{hinton2018matrix}  & ResNet & 10.78 & 10.35 & 7.53 & 4.97 \\
Dynamic Routing \cite{sabour2017dynamic} & ResNet & 10.65 & 8.36 & 7.21 & 5.26 \\
Self Attention Routing \cite{hahn2019self} & ResNet & 10.23 & 8.41 & 6.79 & 3.67 \\ 
Inverted Dot-Product Attention Routing \cite{tsai2020capsules}& ResNet & 9.84 & 8.13 & 5.82 & 3.65 \\\rowcolor{LightCyan} 
Hybrid Gromov-Wasserstein Routing -A & ResNet & 9.93 & 8.56 & 6.74 & 3.81\\\rowcolor{LightCyan} 
Hybrid Gromov-Wasserstein Routing -B & ResNet & 10.62 & 9.41 & 7.57 & 4.12\\\rowcolor{LightCyan} 
Hybrid Gromov-Wasserstein Routing -C & ResNet & 8.68 & 8.27 & 6.91 & 3.74\\\rowcolor{LightCyan} 
Hybrid Gromov-Wasserstein Routing (ours) & ResNet & {\bf 8.32} & {\bf 7.26}& {\bf 5.37} & {\bf 3.21}\\ [0.5ex]
\cline {1-6}
\rule{0pt}{1\normalbaselineskip}ResNet-18 \cite{he2016deep} & & 9.73 & 7.51 & 6.12 & 3.68 \\ 
Baseline CNN \footref{note1} (simple) && 10.62 & 9.27 & 8.31 & 7.54 \\\hline  
\end{tabular}
\end{table*}

\subsection{Computational Cost (Q3)}
\begin{figure}[h!]
\hspace{-4mm}
\includegraphics[height=6.4cm]{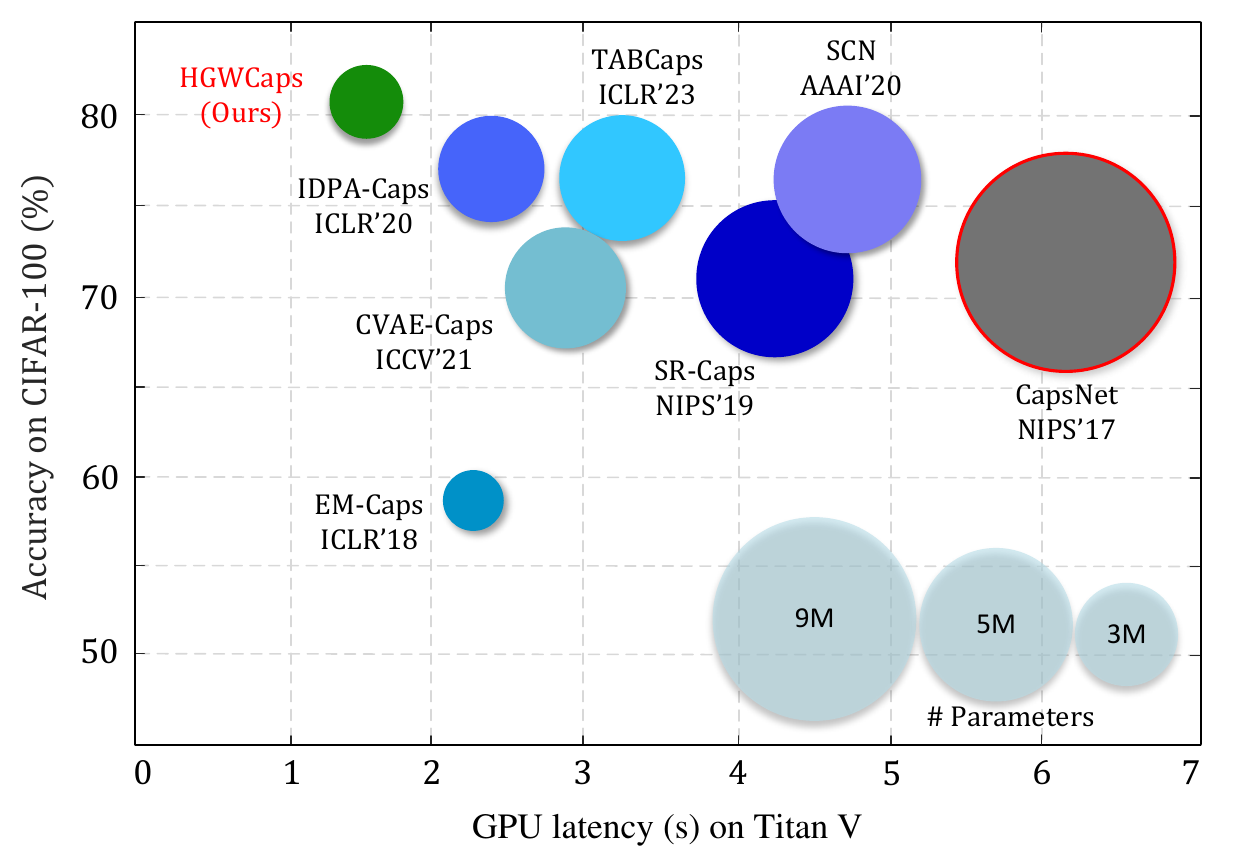} %
\caption{Accuracy, GPU, latency, and parameter comparison on CIFAR-100. The proposed HGWCapsule is 3.2$\times$-11.2$\times$ smaller and 1.8$\times$- 4.5$\times$ faster than other approaches. }   
\label{comp}
\end{figure}

We used the same CNN-based architecture for all CapsNets to have a fair comparison. ResNet-20 \cite{he2016deep} is selected for CIFAR-100 \cite{krizhevsky2009learning} classification since it represents SOTA CNNs for various computer vision applications. Seven strong CapsNets are chosen: CapsNets \cite{sabour2017dynamic}, EM-Capsule \cite{hinton2018matrix}, IDPA-Capsule \cite{tsai2020capsules}, CVAE-Capsule \cite{guo2021conditional}, TABCapsule \cite{chen2022tabcaps}, SR-Capsule \cite{hahn2019self}, and SCN \cite{edraki2020subspace}. Fig. \ref{comp} compares the efficiency of our proposed model with the other baselines on the CIFAR-100 classification task. As we can see, EM-Capsule, which is the lightest capsule version with a ResNet backbone, can only achieve 58.08\% test accuracy, but our model significantly outperforms the corresponding capsule baseline. For the number of parameters, HGWCapsule, EM-Capsule, TABCapsule, and IDPACapsule have much fewer than other baselines due to the different structures of the capsule's pose. Results show that HGWCapsule is up to $2.7\times$ and $2.5\times$ respectively faster than SCN and SR-Capsule. Moreover, HGWCapsule is up to $4.1\times$ and $3.5\times$ respectively smaller than TABCapsule and IDAP-Capsule. We also found that it is hard for some of the CapsNets to reach the same performance as ResNet on the CIFAR-100 dataset, e.g., SR-Capsule and CVAE-Capsule, achieving 68.56\% and 69.12\% test accuracy, while the ResNet backbone archives 77.53\%. 

\subsection{Ablation Study (Q4)}
\begin{figure}
\centering
\includegraphics[height=6.5cm]{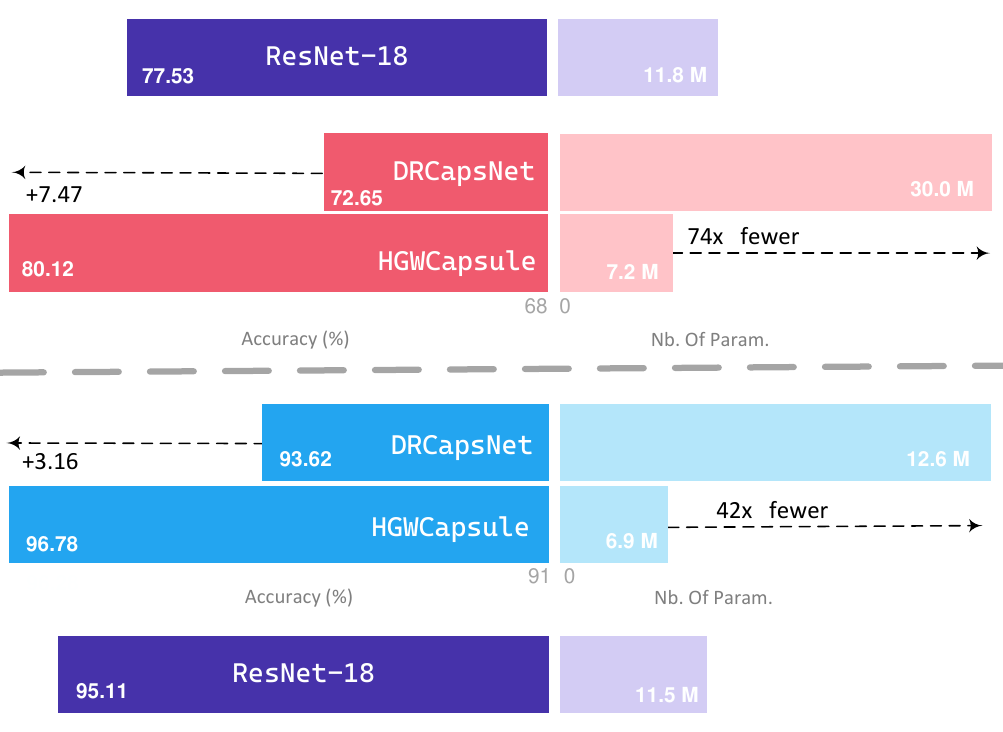}
\caption{We compare the test accuracy of our model with DRCapsNet, utilizing the ResNet-18 backbone for both. The upper results pertain to CIFAR-100, while the lower bars represent CIFAR-10. Remarkably, HGWCapsule achieves superior performance over DRCapsNet, even while employing significantly fewer parameters. This combination of HGWCapsule with ResNet not only reduces parameter complexity but also enhances overall performance. }\label{caps}
\end{figure}

In this section, we discuss our routing mechanism with the following ablations. In addition to our proximal point algorithm, an alternative solution for Eq.(\ref{eq:24}), is to replace the KL-divergence in Eq.(\ref{eq:25}) with an entropy regularizer $H(T)$ (as used in \cite{bunne2019learning}). Therefore, our model baselines are: 1) HGW Routing-A: replacing proximal point algorithm with an entropy regularizer; 2) HGW Routing-B: without adding a regularization term (see Eq. \ref{eq:21}). 3) HGW Routing-C: without using observed data in computing distance matrix $C_\theta(X_\theta)$. Four robust routing approaches are chosen: Dynamic Routing Capsule \cite{sabour2017dynamic}, EM-Routing Capsule \cite{hinton2018matrix}, IDPA-Routing Capsule \cite{tsai2020capsules}, and SA-Routing Capsule \cite{hahn2019self}. Two backbone feature models are used for each CapsNets; standard convolution\footnote{\label{note1} simple CNN is considered as three convolution layers (3$\times$3, stride=2, 2, 1, per layer) followed by a fully-connected layer. The last convolution layer has a pooling layer of size 2$\times$2, stride 2 (same as the work in \cite{sabour2017dynamic})}, and a ResNet-18 \cite{he2016deep}. We consider two modifications while using the ResNet-18; i) the simple feature backbone is replaced with the ResNet feature backbone, and the input dimension is set after the spine to 128.
For the ablation, we use the entire SVHN \cite{netzer2011reading}, since it contains a natural range of geometric variation and is well-suited for this evaluation. The training set has 73,257 image samples; we randomly select 5,000 samples for testing and the remaining 68,257 for training. In addition, the dataset contains 531,131 extra examples that can be used for training. Thus, our training data is performed on 10,000, 30,000, 60,000, and 600,000 examples. 
The results are presented in Table \ref{tab:abl}. In a general trend, our model performs better than the baselines, and IDPA-routing \cite{tsai2020capsules} SA-Routing \cite{hahn2019self} serve better than EM-routing \cite{hinton2018matrix} and DR-routing \cite{sabour2017dynamic}. We also found that, with a simple backbone, it is difficult for Dynamic Routing and EM-Routing to reach the same level of CNNs; e.g., SA-Routing achieves 8.42\% error rates while the baselines CNN archives 7.54\%. The performance of all CapsNets improves when a simple CNN is replaced with the ResNet backbone. This isn't a surprise because ResNet generalizes better than a simple CNN. Our model improves the error rate of the Resent by 14\%, by reducing it from 3.68\% to 3.21\% (we can see a similar observation for simple CNN). 
We also found that when removing the regularizer term, the performance of our model dramatically drops. This result indicates that the regularization step is crucial and can prevent overfitting problems. We notice a performance deterioration when replacing the proximal point KL with an entropy regularizer $H$. Typically, $H(T)$ is more sensitive to the choice of the hyperparameter $\epsilon$ (i.e., it cannot operate for small values of $\epsilon$ ). However, our model is more robust to the variation of $\epsilon$, and we can choose it in a wide range (we set it to 0.001 in our implementation). Moreover, computing the GW distance ($C_\theta (X_\theta)=(1-\beta)C_\theta + \beta  K (X_\theta, X_\theta)$, for $\theta=p$ or $q$) without considering the information of the observed data ($C_\theta$), (means $\beta=1$), it will again degrade the performance. Therefore, both observed and embedded information are necessary. To conclude, combining HGWCapsule with the ResNet backbone gives better results even with limited training data (e.g., 10k). \\
\indent In Fig. \ref{caps}, we also present the test accuracy results of our model compared to DRCapsNet \cite{sabour2017dynamic} on the CIFAR-10 and CIFAR-100. For a fair comparison, both models use ResNet-18 as the backbone. The experimental results show that the HGWCapsNet outperforms the DRCapsNet on both datasets with a few number of parameters. Particularly, the performance gap is more significant on CIFAR-100, where HGWCapsNet achieves a test accuracy of 80.12\% compared to 72.65\% achieved by DRCapsNet. It is worth noting that DRCapsNets faced limitations in handling larger datasets like ImageNet due to memory constraints, making it impractical to perform experiments on that dataset. 

\section{Conclusions}
\label{conc}
We introduced a novel Hybrid Gromov-Wasserstein (HGW) framework for capsule learning, marking a significant advancement for CapsNets in tackling large-scale vision tasks. Similar to the approach in \cite{edraki2020subspace}, we constructed capsules through groups of subcapsules, where each subcapsule modeled potential variations in object parts for a specific class. However, we incorporated the HGW framework to estimate the alignment score between the input and individual subcapsules. This approach diverges from other capsule-based frameworks that typically calculate agreement between capsules using their pose vector dot products. Instead, HGW computes dissimilarity between the input and subcapsules, followed by optimal transport-based correspondence determination. By aligning input and subcapsules in a more structured and optimized manner, HGW enables a more compact data representation, reducing computational load and enhancing overall model efficiency. Empirical validation against various CapsNet and CNN baselines attests to the effectiveness of our proposal.
While HGW shows promise in aligning input and subcapsules through distribution matching and offers efficient routing in CapsNets, its application to unsupervised distribution alignment issues, especially in complex multi-modal data distributions, presents significant challenges. Mainly, when dealing with multi-modal data structures, the alignment process becomes more complex, involving simultaneous identification and alignment of multiple data modes in both domains. In future endeavors, integrating specific constraints into the optimization process could guide HGW to focus on capturing the most pertinent and meaningful data patterns. 


{\footnotesize{
\bibliographystyle{IEEEtran}
\bibliography{IEEEabrv,IEEEexample}
}}


\section*{Appendix}
\label{proof}
{{\noindent \bf Proof of Proposition 1.}} As shown in \cite{peyre2016gromov} and \cite{benamou2015iterative}, the projection operation is equivalent to solving the regularized transport problem. This problem aims to find the optimal transportation plan between two sets, $X_p$ and $X_q$, by minimizing the KL-divergence between them while considering a regularization term to ensure stability. The optimal transport between $X_p$ and $X_q$ is as


\begin{equation}
\begin{array}{lr}
\mathcal{T}(C, \mu_p, \mu_q) = \text{argmin}~ \langle C, T\rangle - \epsilon H(T)
\end{array}
\label{pro}
\end{equation} 

The update step for the projection involves employing projected gradient descent, with both the gradient and projections computed using the KL metric, which can be expressed as 

\begin{equation}
\begin{array}{lr}
T \leftarrow \text{Proj}_{\Gamma(\mu_p,\mu_q)}^{KL} (T \odot e^{- r(\nabla \mathcal{E}_{C_p, C_q} (T) - \epsilon \nabla H(T))}).
\end{array}
\end{equation} 

Here, $r>0$ is a small step size, and $\text{Proj}_{\Gamma(\mu_p,\mu_q)}^{KL} (K) = \text{argmin}~KL (T'|K)$ is the KL projector for any matrix $K$, where $T' \in \Gamma(\mu_p,\mu_q)$.  
However, when $r = 1/\epsilon$, Eq. \ref{pro} can be simplified as $T \leftarrow \mathcal{T}(L(C_p, C_q) \otimes T, \mu_p, \mu_q)$, and the projection operation is precisely the solution to the regularized optimal transport. This is given by: 
$\text{Proj}_{\Gamma(\mu_p,\mu_q)}^{KL} (K) = \mathcal{T}(\epsilon \log(K), \mu_p, \mu_q)$. \\

It also has form of $\nabla \mathcal{E}_{C_p, C_q} (T) - \epsilon \nabla H(T) = L(C_p, C_q) \otimes T + \epsilon \log (T)$.
This formulation leads to a surprisingly simple algorithm, where each update of $T$ involves a Sinkhorn projection. As proved in \cite{benamou2015iterative}, using a learning rate of $1/\epsilon$ leads to a converging sequence of $T$, and this approach works effectively in practice.

\end{document}